\documentclass[11pt, a4paper, logo, copyright]{googlecloud}

\pdftrailerid{redacted}

\makeatletter
\renewcommand\bibentry[1]{\nocitep{#1}{\frenchspacing\@nameuse{BR@r@#1\@extra@b@citeb}}}
\makeatother

\usepackage{kantlipsum, lipsum}
\usepackage{dsfont}

\definecolor{ourred}{HTML}{F19C99}
\definecolor{ourblue}{HTML}{7EA6E0}
\definecolor{plotred}{HTML}{f77189}
\definecolor{plotgreen}{HTML}{33b07a}
\definecolor{plotpurple}{HTML}{cc7af4}
\definecolor{plotmagenta}{HTML}{f565cc}
\definecolor{plotazure}{HTML}{38a9c5}

\definecolor{codegreen}{rgb}{0,0.6,0}
\definecolor{codegray}{rgb}{0.5,0.5,0.5}
\definecolor{codepurple}{rgb}{0.58,0,0.82}
\definecolor{backcolour}{rgb}{0.95,0.95,0.92}

\definecolor{Gray}{gray}{0.92}

\definecolor{stage1}{HTML}{34A853}
\definecolor{stage2}{HTML}{A680B8}
\definecolor{stage3}{HTML}{009999}

\usepackage{listings}

\lstdefinestyle{mystyle}{
    backgroundcolor=\color{backcolour},   
    commentstyle=\color{codegreen},
    keywordstyle=\color{magenta},
    numberstyle=\tiny\color{codegray},
    stringstyle=\color{codepurple},
    basicstyle=\ttfamily\scriptsize,
    breakatwhitespace=false,         
    breaklines=true,                 
    captionpos=b,                    
    keepspaces=true,                 
    numbersep=5pt,                  
    showspaces=false,                
    showstringspaces=false,
    showtabs=false,                  
    tabsize=2,
    frame=none,
    aboveskip=1pt,
    belowskip=1pt,
}
\usepackage[most]{tcolorbox}
\tcbset{enhanced,
boxrule=0pt,frame hidden,
borderline west={4pt}{0pt}{green!75!black},
top=0pt,bottom=0pt,
colback=green!10!white,
sharp corners}

\usepackage{colortbl}
\usepackage{float}

\usepackage{multirow}

\usepackage[authoryear, sort&compress, round]{natbib}

\usepackage[utf8]{inputenc} %
\usepackage[T1]{fontenc}    %
\usepackage{hyperref}       %
\hypersetup{
  colorlinks,
  citecolor=blue,
  linkcolor=red,
  urlcolor=blue}
\usepackage{url}            %
\usepackage{booktabs}       %
\usepackage{amsfonts}       %
\usepackage{nicefrac}       %
\usepackage{microtype}      %
\usepackage{wrapfig}        %
\usepackage{graphicx}       %
\usepackage{soul}           %
\usepackage{enumitem}
\usepackage{amssymb,amsmath}
\usepackage{tikz, adjustbox}
\usepackage[most]{tcolorbox}
\usepackage{subcaption}
\usepackage{multirow}
\usepackage{arydshln}
\usepackage{float}
\usepackage{algorithm}
\usepackage{algorithmic}
\usepackage[capitalize,noabbrev]{cleveref}
\usepackage{fancyvrb}
\usepackage{tcolorbox}
\usepackage{xspace}
\usepackage{lineno}
\usepackage{titlesec}
\usepackage{textcomp}
\usepackage{pifont}
\usepackage[table]{xcolor}
\usepackage{fontawesome5}
\usepackage{tabularx}
\usepackage{makecell}
\usepackage{listings}

\usepackage{amsmath,amsfonts,bm}

\def\eqref#1{equation~\ref{#1}}

\def\1{\bm{1}}

\DeclareMathAlphabet{\mathsfit}{\encodingdefault}{\sfdefault}{m}{sl}
\SetMathAlphabet{\mathsfit}{bold}{\encodingdefault}{\sfdefault}{bx}{n}

\titlespacing{\paragraph}{%
  0pt}{
  0pt}{
  1em}

\newcommand{\logo}{\raisebox{-6pt}{\includegraphics[width=2em]{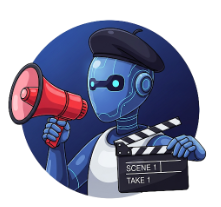}}}

\newcommand{\moniker}{\textsc{Co-Director}\xspace}
\newcommand{\benchmark}{\textsc{GenAd-Bench}\xspace}

\newcommand{\copyrightext}{}

\renewcommand{\today}{}

\definecolor{darkblue}{rgb}{0, 0, 0.5}
\hypersetup{colorlinks=true, citecolor=darkblue, linkcolor=darkblue, urlcolor=darkblue}

\lstdefinestyle{mystyle}{
    backgroundcolor=\color{backcolour},   
    commentstyle=\color{codegreen},
    keywordstyle=\color{magenta},
    numberstyle=\tiny\color{codegray},
    stringstyle=\color{codepurple},
    basicstyle=\ttfamily\scriptsize,
    breakatwhitespace=false,         
    breaklines=true,                 
    captionpos=b,                    
    keepspaces=true,                 
    numbers=left,                    
    numbersep=5pt,                  
    showspaces=false,                
    showstringspaces=false,
    showtabs=false,                  
    tabsize=2,
    frame=none,
    aboveskip=1pt,
    belowskip=1pt,
}
\lstdefinestyle{plainins}{
    backgroundcolor=\color{white},   
    commentstyle=\color{codegreen},
    keywordstyle=\color{magenta},
    numberstyle=\tiny\color{codegray},
    stringstyle=\color{codepurple},
    basicstyle=\ttfamily\scriptsize,
    breakatwhitespace=false,         
    breaklines=true,                 
    captionpos=b,                    
    keepspaces=true,                 
    numbers=none,                    
    numbersep=5pt,                  
    showspaces=false,                
    showstringspaces=false,
    showtabs=false,                  
    tabsize=2,
    aboveskip=0pt,
    belowskip=0pt,
    frame=single
}
\lstdefinestyle{plainexam}{
    backgroundcolor=\color[HTML]{FFFCF3},   
    commentstyle=\color{codegreen},
    keywordstyle=\color{magenta},
    numberstyle=\tiny\color{codegray},
    stringstyle=\color{codepurple},
    basicstyle=\ttfamily\scriptsize,
    breakatwhitespace=false,         
    breaklines=true,                 
    captionpos=b,                    
    keepspaces=true,                 
    numbers=none,                    
    numbersep=5pt,                  
    showspaces=false,                
    showstringspaces=false,
    showtabs=false,                  
    tabsize=2,
    aboveskip=0pt,
    belowskip=0pt
}

\title{\logo{} \moniker: Agentic Generative Video Storytelling}

\author[1 *]{Yale Song}
\author[1]{Yiwen Song}
\author[1]{Nick Losier}
\author[1]{Nathan Hodson}
\author[1]{Ye Jin}
\author[1]{Rhyard Zhu}
\author[1]{Yan Xu}
\author[1]{Daniel Vlasic}
\author[1]{Carina Claassen}
\author[1]{Jasmine Leon}
\author[1]{Khanh G. LeViet}
\author[1]{Zack Chomyn}
\author[1]{Joe Timmons}
\author[1]{Brett Slatkin}
\author[1]{Scott Penberthy}
\author[1]{Tomas Pfister}
\affil{Google Inc.}
\correspondingauthor{yalesong@google.com}

\begin{abstract}
While diffusion models generate high-fidelity video clips, transforming them into coherent storytelling engines remains challenging. Current agentic pipelines automate this via chained modules but suffer from semantic drift and cascading failures due to independent, handcrafted prompting. We present \moniker, a hierarchical multi-agent framework formalizing video storytelling as a global optimization problem. To ensure semantic coherence, we introduce hierarchical parameterization: a multi-armed bandit globally identifies promising creative directions, while a local multimodal self-refinement loop mitigates identity drift and ensures sequence-level consistency. This balances the exploration of novel narrative strategies with the exploitation of effective creative configurations. For evaluation, we introduce \benchmark, a 400-scenario dataset of fictional products for personalized advertising. Experiments demonstrate that \moniker significantly outperforms state-of-the-art baselines, offering a principled approach that seamlessly generalizes to broader cinematic narratives.\\ \\Project Page: \url{https://co-director-agent.github.io/}
\end{abstract}

\begin{document}

\maketitle

\section{Introduction}
\label{sec:intro}

Generative video storytelling aims to transform abstract ideas---such as a simple tagline and  sparse reference visuals---into cohesive video narratives, like a fully realized short film. It has diverse applications, ranging from democratizing high-end content creation~\citep{divya2024transforming} to automating filmmaking~\citep{zhang2025generative} and digital advertising~\citep{baek2023digital}. Realizing this potential requires jointly optimizing multiple objectives: narrative coherence, visual consistency, and \textit{mise-en-scène}---the visual language dictating how camera angles, color palettes, and metaphors deliver the intended message~\citep{zettl1973sight}.

Recent advancements in video diffusion demonstrate remarkable high-fidelity generation. High-profile collaborations, such as Google DeepMind's work with Darren Aronofsky on \textit{Ancestra}~\citep{ancestra2025} and \textit{On This Day... 1776}~\citep{onthisday2026}, showcase the artistic ceiling when backed by a 200-person creative team. While representing the pinnacle of human-AI collaboration, such resources remain out of reach for independent creators. Without them, individuals must manually bridge the gap between LLM scripting and fragmented generative models, correcting for the consistency errors and identity drift inherent in current architectures~\citep{elmoghany2025survey}.

To mitigate this, the community has pivoted toward agentic pipelines where LLMs orchestrate the multi-step generative process~\citep{lin2023videodirectorgpt,shi2025animaker,wu2025automated,wei2025hollywood,huang2025filmaster}. However, these systems remain susceptible to ``cascading failures,'' where early errors propagate and break long-horizon consistency~\citep{zhou2024storydiffusion}. This reflects the classical credit assignment problem~\citep{minsky1961steps}, as terminal failures are difficult to trace back to specific prompts. Furthermore, existing systems rely on static, handcrafted prompt templates lacking a unified creative vision. Without a mechanism to map script-level changes to downstream generation, the resulting visual output often feels disjointed. Crucially, these approaches fail to capture the \textit{serendipity} essential to creative production. Human directors often explore counter-intuitive ideas to find ``creative gold.'' Static, chained pipelines are fundamentally ill-equipped for such exploration, remaining anchored to fixed templates unable to adapt to the non-linear nature of creative success.

In this work, we present \moniker, a multi-agent framework for automating video storytelling. To the best of our knowledge, \moniker\ is the first system to formalize generative storytelling as a global optimization problem. Unlike prior agentic pipelines that rely on linear, "waterfall" chains prone to semantic drift, we introduce a \textit{hierarchical parameterization} strategy steered by Multi-Armed Bandit (MAB)~\citep{bubeck2012regret}. Rather than relying on fixed prompt templates, \moniker utilizes MAB to sample abstract creative trajectories and dynamically injects these into the system prompts of sub-agents. This top-down steering ensures that all-agents---from script-writing to video generation---are orthogonally aligned toward a unified creative intent. This formulation formalizes the creative process as a search for the optimal balance between \textit{exploration}---venturing into novel, unconventional narrative territories---and \textit{exploitation}---refining historically effective stylistic configurations. This allows \moniker\ to move beyond simple self-correction toward a principled exploration of the creative latent space, identifying the configuration that optimally balances visual fidelity with narrative effectiveness~\citep{escalas2004narrative}.

Furthermore, we argue that evaluating video storytelling requires a shift from open-ended cinematic metrics towards rigorous, constraint-driven frameworks. We use digital advertising as our demonstration scenario because it demands both sophisticated storytelling (much like short cinematic film) and strict adherence to objective requirements: asset preservation, audience alignment, and value proposition communication~\citep{teixeira2012emotion}. To this end, we introduce \benchmark, a dataset of 400 scenarios across 200 fictional products. By utilizing fictional entities, we respect copyright constraints while preventing models from defaulting to memorized training priors. We provide a multimodal LLM (MLLM)-based evaluation suite covering \textit{Visual Asset Fidelity}, \textit{Demographic Alignment}, \textit{Visual Quality}, and \textit{Marketing Appeal}---anchored in the AIDA hierarchy~\citep{strong1925psychology, barry1990review}.

Our primary contributions are summarized as follows:
\begin{itemize}[leftmargin=20pt, itemsep=3pt, parsep=0pt, topsep=1pt]
\item We propose \moniker, a multi-agent architecture for video storytelling that achieves global coherence by integrating multi-armed bandit optimization with a local, MLLM feedback-driven refinement loop.
\item We introduce the \textit{hierarchical parameterization} of creative intent, enabling the consistent propagation of visual and narrative styles across discrete generative sub-agents.
\item We present \benchmark to rigorously evaluate generative video storytelling across 400 scenarios, eliminating memorization bias via the use of fictional brand assets.
\item We demonstrate our approach significantly outperforming existing foundation models and agentic baselines in asset preservation, contextual alignment, and visual quality.
\item We open-source our implementation of \moniker and the \benchmark dataset to facilitate future research in generative video storytelling.
\end{itemize}

\section{Related Work}

\paragraph{Foundation Models for Video Synthesis}
While large-scale diffusion models~\citep{veo3.1, bytedance2026seedance, kling2026kling3omni, wan2025wan} achieve unprecedented cinematic realism, maintaining long-range consistency remains challenging~\citep{elmoghany2025survey}. Specialized architectures like Phantom~\citep{liu2025phantom} and MoC~\citep{cai2025mixture} introduce advanced techniques to maintain character consistency, but they remain fundamentally standalone generators. Furthermore, these models serve as powerful building blocks rather than complete end-to-end solutions, lacking the high-level planning and reasoning required to create complex narratives in the resulting video output~\citep{elmoghany2025survey}.

\paragraph{Agentic Frameworks for Video Storytelling}
To bridge the gap between clip generation and narrative production, recent multi-agent systems utilize LLMs to decompose scripts into shot lists and prompts~\citep{lin2023videodirectorgpt, wu2025automated}, targeting domains like personalized vlogs~\citep{hou2025personavlog}, music videos~\citep{tang2025automv}, and short films~\citep{zhao2024moviedreamer, xiao2025captain, wei2025hollywood, mu2026script}. However, these ``waterfall'' pipelines are susceptible to cascading errors. While frameworks like AniMaker~\citep{shi2025animaker} and MAViS~\citep{wang2026mavis} incorporate local consistency checks, they struggle with global credit assignment~\citep{minsky1961steps} across the agentic stack. To resolve this, \moniker\ moves beyond the local-refinement paradigm by introducing a Multi-Armed Bandit (MAB) formulation~\citep{lai1985asymptotically, bubeck2012regret} to perform a principled exploration-exploitation of the creative latent space. This global, hierarchical parameterization resolves the ``semantic collisions'' between sub-agents that plague current pipelines, preventing identity drift and ensuring horizontal cohesiveness.

\paragraph{Benchmarks for Video Storytelling}
As generative capabilities evolve, evaluation methodologies must adapt to measure higher-order reasoning. Generic video benchmarks~\citep{huang2024vbench,sun2025t2v} focus on prompt alignment and broad aesthetics, lacking the visual and narrative qualities demanded by professional workflows. Furthermore, even recent narrative-focused benchmarks, such as MovieBench~\citep{wu2025moviebench} and ViStoryBench~\citep{zhuang2025vistorybench}, strictly evaluate isolated generative sub-tasks using meticulously pre-defined scripts. Consequently, current literature lacks evaluations for true end-to-end storytelling: the complex process of autonomously transforming a simple, abstract idea into a cohesive video narrative. Our work addresses this gap by introducing \benchmark designed specifically for end-to-end generative workflows.

\begin{figure}
    \centering 
    \includegraphics[width=\columnwidth]{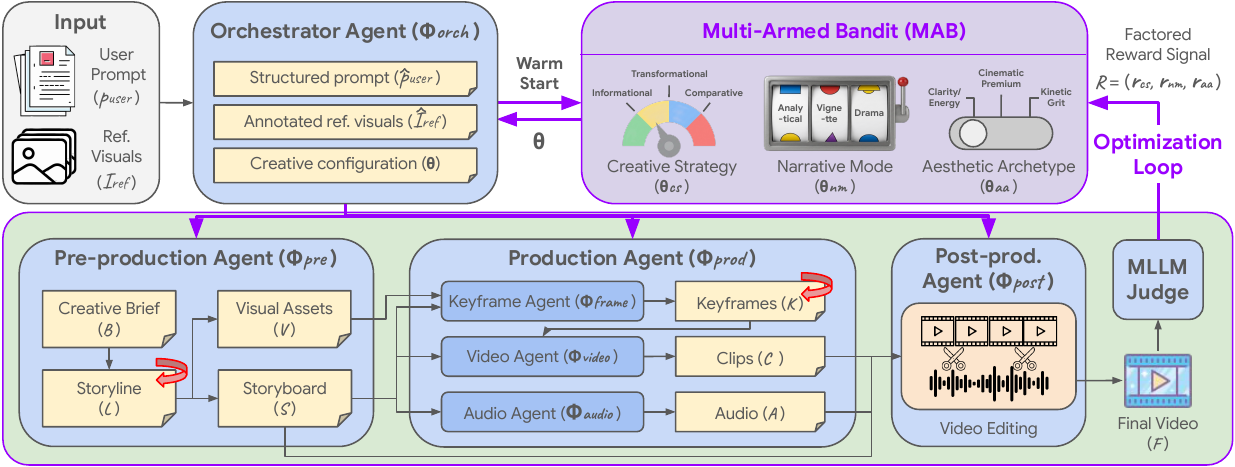}
    \caption{\textbf{\moniker} multi-agent pipeline: 
    \textbf{(Top):} The Orchestrator Agent ($\Phi_{orch}$) utilizes MAB to navigate a factored creative action space $\theta = (\theta_{cs}, \theta_{nm}, \theta_{aa})$. 
    \textbf{(Bottom):} The Pre-Production Agent ($\Phi_{pre}$) synthesizes a creative brief ($\mathcal{B}$), storyline ($\mathcal{L}$), and visual assets ($\mathcal{V}$) to establish a consistent storyboard ($\mathcal{S}$), which the Production Agent ($\Phi_{prod}$) translates into synchronized keyframes ($\mathcal{K}$), video clips ($\mathcal{C}$), and audio ($\mathcal{A}$). An MLLM Judge evaluates the final video ($F$) to generate a factored reward signal $R = (r_{cs}, r_{nm}, r_{aa})$, which is propagated back to the MAB to iteratively refine the creative configuration over $T$ optimization loops.}
    \label{fig:pipeline}
\end{figure}

\section{Approach}
\subsection{Hierarchical Multi-Agent Architecture}
\label{sec:mas_architecture}
We formulate \moniker as a hierarchical multi-agent system orchestrating generative video workflows. While the framework is domain-agnostic, we instantiate it to solve \textit{video advertising}, which comes with strict narrative and visual constraints. 

\paragraph{Orchestrator Agent ($\Phi_{orch}$)} 
This agent manages the global state and execution flow of all sub-agents. Before downstream generation, it performs multimodal data ingestion: 1) parsing the user prompt $p_{user}$ into structured constraints $\hat{p}_{user}$ (brand, product, and demographics~\citep{baardman2021dynamic}); 2) annotating reference visuals $\mathcal{I}_{ref}$ with captions and semantic roles (e.g., product, logo, protagonist), yielding $\hat{\mathcal{I}}_{ref}$. Post-initialization, $\Phi_{orch}$ drives a global optimization loop for $T$ iterations. At iteration $t$, it interfaces with a multi-armed bandit (Section~\ref{sec:approach_mab}) to sample a creative configuration $\theta^{(t)}$, forming a global conditioning context $\Theta^{(t)} = (\hat{p}_{user}, \hat{\mathcal{I}}_{ref}, \theta^{(t)})$,\footnote{We omit superscript $(t)$ hereafter for clarity.} which is then propagated downstream.

\paragraph{Pre-Production Agent ($\Phi_{pre}$)} 
This agent transforms the structured constraints $\Theta$ into a descriptive storyboard $\mathcal{S}$ through four sequential modules:
\begin{itemize}[leftmargin=15pt, itemsep=5pt, parsep=0pt, topsep=1pt]

    \item \textbf{Creative Brief ($\mathcal{B}$):} Acting as the ``deep research'' phase, this module performs contextual enrichment, expanding  product/demographic constraints within $\hat{p}_{user}$ into localized cultural resonance and environmental context. This becomes a \textit{creative brief} $\mathcal{B}$.

    \item \textbf{Storyline ($\mathcal{L}$):} Conditioned on $\mathcal{B}$ and $\theta$, this module constructs the \textit{storyline} $\mathcal{L}$ that acts as a scene-by-scene script establishing plot, character actions, pacing, and transitions. 

    \item \textbf{Visual Assets ($\mathcal{V}$):} This module generates \textit{visual assets} $\mathcal{V}$ for entities in $\mathcal{L}$ (e.g., characters, props, environments) only if they are absent from the reference visuals $\hat{\mathcal{I}}_{ref} \in \Theta$. For characters, it synthesizes context-aligned wardrobes dictated by $\mathcal{L}$, compositing the base character and garments into multi-angle reference collages.

    \item \textbf{Storyboard ($\mathcal{S}$):} Expands $\mathcal{L}$ into a storyboard $\mathcal{S} = \{s_0, \dots, s_{N-1}\}$ modulated by $\theta$. Each scene $s_i \in \mathcal{S}$ defines \textit{semantic descriptors} (actions, environment), \textit{camera kinematics} (scale, movements), \textit{temporal constraints}, and \textit{entity presence flags} linked to $\mathcal{V}$. It also formulates global audio directives (voiceover, tempo, mood), inspired by consumer psychology literature~\citep{gorn1982effects,alpert1990music}.
    
\end{itemize}

\paragraph{Production Agent ($\Phi_{prod}$)} 
This agent translates the storyboard $\mathcal{S}$ into multimodal artifacts via three specialized sub-agents:
\begin{itemize}[leftmargin=15pt, itemsep=5pt, parsep=0pt, topsep=1pt]

    \item \textbf{Keyframe Agent ($\Phi_{frame}$):} Direct text-to-video generation is prone to identity drift~\citep{elmoghany2025survey}. To mitigate this, $\Phi_{frame}$ establishes a visual prior by generating first-frame images to condition downstream synthesis, modulated by aesthetic constraints in $\theta$ (Section~\ref{sec:approach_mab}). Formally, for each scene, it processes descriptors $s_i$, assigned assets $\mathcal{V}$, and configuration $\theta$ to output the keyframe set $\mathcal{K}$:
    \begin{equation}
        \mathcal{K} = \{k_0, k_1, \dots, k_{N-1}\} \quad \text{where} \quad k_i = \Phi_{frame}(s_i, \mathcal{V}, \theta)
    \end{equation}
    
    \item \textbf{Video Agent ($\Phi_{video}$):} This sub-agent transforms $\mathcal{K}$ into a sequence of video clips $\mathcal{C}$. For each scene, it conditions a video diffusion model on the initial keyframe $k_i$, parameterized by local descriptors $s_i$ and creative configuration $\theta$ to dictate temporal dynamics like pacing and camera kinematics (Section~\ref{sec:approach_mab}). Formally, this yields the clip set $\mathcal{C}$:
        \begin{equation}
        \mathcal{C} = \{c_0, c_1, \dots, c_{N-1}\} \quad \text{where} \quad c_i = \Phi_{video}(s_i, k_i, \theta)
    \end{equation}

    \item \textbf{Audio Agent ($\Phi_{audio}$):} This sub-agent synthesizes the acoustic elements specified in $\mathcal{S}$. Rather than per-scene generation, it processes global voiceover and background music from the planning phase. Acoustic properties (e.g., tempo, genre, mood) are conditioned on $\theta$. $\Phi_{audio}$ employs text-to-speech and text-to-audio models to generate and mix these discrete elements into a unified global track $\mathcal{A}$. Formally:
    \begin{equation}
        \mathcal{A} = \Phi_{audio}(\mathcal{S}, \theta)
    \end{equation}

\end{itemize}

\paragraph{Post-Production Agent ($\Phi_{post}$)} 
This module consolidates clips $\mathcal{C}$ and audio $\mathcal{A}$ into the final video $F$. It concatenates $\mathcal{C}$ based on temporal constraints defined in storyboard $\mathcal{S}$, then multiplexes $\mathcal{A}$ onto the sequence to synchronize visual pacing with acoustic cues.
\begin{equation}
    F = \Phi_{post}(\mathcal{S}, \mathcal{C}, \mathcal{A})
\end{equation}

\subsection{Local Optimization via Agentic Self-Refinement}
\label{sec:approach_refine}

In sequential generative pipelines, early errors propagate and amplify downstream. To prevent this, \moniker implements \textit{feedback descent}~\citep{yuksekgonul2024textgrad,lee2025feedback} to self-refine intermediate artifacts. LLMs and MLLMs iteratively evaluate and refine artifacts based on textual feedback. This loop terminates upon reaching a score threshold or iteration budget, returning the best attempt. We apply this \textit{local optimization} to storyline $\mathcal{L}$ and keyframes $\mathcal{K}$, reserving final video evaluation for a global reward signal (Section~\ref{sec:approach_mab})

\paragraph{Storyline}
Before $\Phi_{pre}$ generates storyboard $\mathcal{S}$, an LLM assesses storyline $\mathcal{L}$ on hook quality, narrative cohesion, product integration, emotional engagement, and prompt adherence. It outputs a score with textual feedback and rewrites the prompt to emphasize areas needing improvement. If the score falls below threshold $\tau_{\mathcal{L}}$, the agent regenerates $\mathcal{L}$ using this revised prompt, ensuring a coherent narrative foundation before expansion into $\mathcal{S}$.

\paragraph{Keyframe}
An MLLM evaluates the keyframe sequence $\mathcal{K} = \{k_0, \dots, k_{N-1}\}$ jointly rather than individually. This \textit{joint contextualization} evaluates sequence-level consistency for character and product identities, alongside environmental continuity. Comparing $\mathcal{K}$ against $\hat{p}_{user}$ and $\mathcal{V}$, it scores visual consistency, narrative flow, product appeal, and prompt adherence. If the aggregate score falls below threshold $\tau_{\mathcal{K}}$, the MLLM isolates issues and indexes problematic keyframes in $\mathcal{K}$. It then provides actionable feedback and refined prompts, allowing $\Phi_{frame}$ to selectively regenerate flagged keyframes while preserving accepted ones.

\subsection{Global Optimization via Multi-Armed Bandits}
\label{sec:approach_mab}

Because the chained pipeline is non-differentiable, standard gradient updates cannot achieve end-to-end optimization. Instead, \moniker uses \textit{hierarchical parameterization}, treating the pipeline as a black box steered by centralized parameters governing global creative directions. We formulate tuning these parameters as a Multi-Armed Bandit (MAB) problem~\citep{lai1985asymptotically,bubeck2012regret}. Operating at this top level, rather than tuning individual prompts independently, ensures semantic coherence across sub-agents while balancing the exploration of novel strategies with the exploitation of proven configurations.

\paragraph{MAB Preliminaries}
In a standard MAB framework, an agent iteratively selects an action (an ``arm''), observes a reward, and updates its policy to maximize long-term payout. The central challenge is the \textit{exploration-exploitation tradeoff}: balancing testing untried arms for potentially higher rewards against exploiting the best-known option. In \moniker, a single ``pull'' corresponds to a full pipeline execution, the ``arms'' are creative directions $\theta$, and the ``reward'' is the terminal video evaluation score.

\paragraph{Hierarchical Parameterization}
We define the action space (the ``arms'') over three orthogonal axes of video production grounded in marketing and media theory. The creative direction  $\theta$ is a tuple comprising three categorical variables: $\theta = (\theta_{cs}, \theta_{nm}, \theta_{aa})$.
\begin{itemize}[leftmargin=15pt, itemsep=5pt, parsep=0pt, topsep=1pt]
    \item \textbf{Creative Strategy ($\theta_{cs}$):} This controls \textit{what} the ad is fundamentally trying to communicate~\citep{laskey1989typology}. The bandit selects among an \textit{Informational} strategy (focusing on functional utility), a \textit{Transformational} strategy (driving psychological and lifestyle appeal), or a \textit{Comparative} strategy (positioning against a known standard or competitor).
    
    \item \textbf{Narrative Mode ($\theta_{nm}$):} This controls \textit{how} the strategy is delivered over time~\citep{escalas2004narrative}. The bandit selects among an \textit{Analytical} mode (an argument-based delivery without a story arc), a \textit{Vignette} mode (atmospheric slices-of-life prioritizing mood over plot), or a \textit{Narrative Drama} mode (a character-driven sequence of conflict and resolution).
    
    \item \textbf{Aesthetic Archetype ($\theta_{aa}$):} This controls the \textit{sensory realization} of the video~\citep{zettl1973sight} via four configurations: \textit{Clarity/Energy} (high-key lighting, fast-cut motion, high-tempo audio); \textit{Cinematic Premium} (chiaroscuro lighting, slow reflective camera work, orchestral audio); \textit{Minimalist Focus} (bright backgrounds, micro-movements, ASMR-style foley); and \textit{Kinetic Grit} (low-key lighting, handheld/FPV drone motion, electronic synth audio).
\end{itemize}
At execution step $t$, the MAB selects an arm combination $\theta^{(t)}$. While downstream LLMs can interpret these creative parameters, independent interpretation risks semantic drift. To ensure coherence, orchestrator $\Phi_{orch}$ centrally synthesizes $\theta^{(t)}$ into agent-specific instructions, dynamically injecting them into placeholders within the agents' system prompts.

\paragraph{Reward}
To evaluate $\theta^{(t)}$, \moniker uses an MLLM evaluator processing the final video $F^{(t)}$ alongside $p_{user}$ and $\mathcal{I}_{ref}$ to yield two metrics (full prompt in Appendix~\ref{app:video_verifier_rubrics}). First, an aggregate quality score (temporal coherence, visual quality, consistency) selects the best video across $T$ executions. Second, it computes dimension-specific \textit{strategic efficacy} scores crucial for MAB optimization. This decouples execution from strategy, assessing if the creative direction was effective independent of downstream diffusion artifacts. This yields a factored reward signal $R^{(t)} = (r_{cs}, r_{nm}, r_{aa}) \in [0, 100]^3$ to update the bandit's policy.

\paragraph{Optimization Policy and Warm Start}
\moniker models the search using the Upper Confidence Bound (UCB1) algorithm~\citep{auer2002finite}, which uses the disentangled scalars from $R^{(t)}$ to independently update the expected value of each selected arm. By decoupling the reward updates, the MAB analytically bypasses the credit assignment problem~\citep{minsky1961steps}, isolating exactly which creative strategies drove the video's success or failure. Furthermore, because unguided exploration is prohibitively expensive in video generation, we introduce an LLM-driven \textit{warm start}. Prior to the first execution, an LLM analyzes the user's demographic and product domain to initialize the expected values of the arms, safely biasing early exploration toward theoretically sound configurations.

\section{\benchmark: Evaluating Generative Video Advertising}

To construct \benchmark, we developed a human-in-the-loop pipeline using Gemini 3 Pro~\citep{gemini3p} for text synthesis and Nano Banana Pro~\citep{gemini3p_image} for visual assets. We generated 50 fictional brands with four products each, spanning consumer goods to industrial equipment. For each product, we authored two contrasting demographics---\textit{stereotypical} and \textit{unconventional}---yielding 400 unique scenarios defined by a six-point prompt (brand, product, gender, age, location, interest). Accompanying visual assets (logos, product references) were generated and manually verified (15\% required regeneration) to ensure uniqueness and prevent visual memorization or copyright conflicts.

The resulting benchmark enforces \textbf{strict targeting constraints} and \textbf{multimodal conditioning}, requiring models to synthesize textual demographic intent with specific reference imagery. By evaluating paired scenarios across a \textbf{global geographic distribution}, \benchmark\ explicitly tests conceptual flexibility and exposure to ingrained biases. Furthermore, by utilizing \textbf{fictional entities}, the benchmark prevents models from defaulting to memorized training priors, ensuring that performance is a measure of reasoning rather than retrieval. We partition the data into half (200 scenarios each) for hillclimbing and validation.

We use Gemini 3 Pro~\citep{gemini3p} as an MLLM-as-a-Judge to evaluate generated videos; see Appendix~\ref{app:rubrics} for the rubric prompts. The MLLM simultaneously processes visual frames and the audio track to score four dimensions (0--100):
\begin{itemize}[leftmargin=12pt, itemsep=3pt, parsep=0pt, topsep=1pt]
    \item \textbf{Visual Asset Fidelity (VAF)}: Measures faithfulness to user-provided logos and products. The MLLM assesses visual-semantic similarity, ensuring core identity preservation across complex camera angles, occlusions, and dynamic lighting.
    
    \item \textbf{Demographic Alignment (DA)}: Assesses targeting success across four variables (\textit{Gender, Age, Location, Interest}). The MLLM penalizes generic aesthetics while rewarding tailored casting, localized environmental context, and narrative tone.
    
    \item \textbf{Marketing Appeal (MA)}: Evaluates advertisement persuasiveness via the AIDA hierarchy~\citep{strong1925psychology,barry1990review} and engagement theory~\citep{teixeira2012emotion}. It scores visual hooks, value proposition clarity, and emotional resonance.
    
    \item \textbf{Visual Quality (VQ)}: A prompt-agnostic metric scanning for diffusion artifacts, object morphing, and temporal flickering. It ensures videos meet broadcast viability by penalizing unnatural motion or violations of basic physical laws.
\end{itemize}
While utilizing an MLLM for evaluation raises potential bias concerns, we benchmark these metrics against a human study (Section~\ref{sec:human_validation}). We find that the MLLM serves as a reliable proxy, showing alignment that approaches the human-human agreement ceiling.

\section{Experiments}
\label{sec:experiments}
We evaluate \moniker on \benchmark against state-of-the-art baselines, demonstrating its superior storytelling capabilities, followed by ablation studies validating the necessity of local self-refinement and global MAB optimization.

\paragraph{Implementation Details} We instantiate \moniker with Gemini 3 Pro~\citep{gemini3p}, Nano Banana Pro~\citep{gemini3p_image}, Veo 3.1~\citep{veo3.1}, Gemini 2.5 Pro TTS~\citep{gemini2.5p_tts}, and Lyria 2~\citep{lyria2}. MAB optimization is set to $T=4$ iterations (see App.~\ref{app:mab_efficiency} for search efficiency trade-offs), and local refinement is capped at 3 retries. Final videos feature 4 shots totaling 12 seconds.

\paragraph{Baselines} We evaluate \moniker against monolithic models, commercial platforms, and multi-agent frameworks, listed in Table~\ref{tab:main_results}. To isolate architectural efficacy, we compare against analogous agentic pipelines \textbf{AniMaker} and \textbf{MovieAgent} using the same underlying models. Since these require scene-by-scene scripts, \textsc{Gemini 3 Pro} expands our six-point prompts into the necessary formats. Finally, we compare \moniker against a \textbf{Base Agentic Pipeline} (no optimizations) and a \textbf{Random Search Baseline ($T=4$)} that samples the creative space without MAB or warm-up. Baseline details are in Appendix~\ref{app:baseline}.

\begin{table*}[tp]
\centering
\small
\caption{\textbf{Evaluation on \benchmark.} All metrics are scaled $[0, 100]$. \textbf{VAF}: Visual Asset Fidelity, \textbf{DA}: Demographic Alignment, \textbf{MA}: Marketing Appeal, \textbf{VQ}: Visual Quality.}
\label{tab:main_results}
\begin{tabular}{lccccc}
\toprule
\textbf{Method}  & \textbf{VAF} $\uparrow$ & \textbf{DA} $\uparrow$ & \textbf{MA} $\uparrow$ & \textbf{VQ} $\uparrow$ & \textbf{Avg.}  $\uparrow$\\
\midrule
\multicolumn{6}{l}{\textit{Proprietary Models}} \\
Creatify~\citep{creatify2026}  & 23.2 & 16.2 & 19.5 & 29.5 & 22.1 \\
HeyGen~\citep{heygen2026}  & 42.9 & 59.3 & 39.5 & 45.0 & 46.7 \\
Kling 3.0 Omni~\citep{kling2026kling3omni} &  62.0 & 70.3 & 56.0 & 45.3 & 58.4 \\
Veo 3.1~\citep{veo3.1} & 60.0 & 80.8 & 63.2 & 50.5 & 63.6 \\
Wan 2.6~\citep{wan2025wan} & 67.0 & 71.5 & 62.5 & 58.9 & 65.0 \\
\midrule
\multicolumn{6}{l}{\textit{Open-Source Models}} \\
LTX-2.3~\citep{hacohen2024ltx} & 23.5 & 56.0 & 30.6 & 24.4 & 33.6 \\
AniMaker~\citep{shi2025animaker} & 53.1 & 81.3 & 60.3 & 53.9 & 62.2 \\
MovieAgent~\citep{wu2025automated} & 61.2 & 81.3 & 66.4 & 52.4 & 65.3 \\
\midrule
Base Agentic Pipeline ($T=1$)& 68.5 & 78.4 & 67.1 & 59.9 & 68.5 \\
Random Search Baseline ($T=4$) & 77.0 & 85.8 & 75.6 & 64.1 & 75.7 \\
\rowcolor{gray!10} \textbf{\moniker ($T=4$)} & \textbf{82.1} & \textbf{91.4} & \textbf{82.0} & \textbf{70.2} & \textbf{81.4} \\
\bottomrule
\end{tabular}
\end{table*}

\subsection{Main Results}
\label{sec:main_results}

As reported in Table~\ref{tab:main_results}, \moniker achieves the highest average performance (81.4). We include specialized commercial platforms (Creatify, HeyGen) to establish the current industry baseline; their reliance on "talking head" avatar synthesis with static overlays fundamentally limits their performance on complex narratives, highlighting the necessity of \moniker's dynamic storytelling. Crucially, the performance gap between \moniker and prior agentic baselines (AniMaker, MovieAgent) stems strictly from agentic architectural differences. By utilizing the same LLM (Gemini 3 Pro) to generate their requisite input scripts, we ensured a highly competitive and equitable starting point. Furthermore, self-refinement (Section~\ref{sec:approach_refine}) and MAB optimization (Section~\ref{sec:approach_mab}) drive \moniker's superior Visual Quality (VQ: 70.2) over its base model, Veo 3.1 (VQ: 50.5). Although VQ is prompt-agnostic, \moniker's optimization proactively improves suboptimal clips, effectively elevating the foundation model's baseline and ensuring superior temporal consistency and physical realism over raw, single-pass generation.

To isolate the impact of our orchestration and global optimization, we evaluate \moniker\ against both external baselines and internal architectural variants. Using identical foundation models, our Base Agentic Pipeline ($T=1$) averages 68.5. This proves our core orchestration alone significantly outperforms prior linear open-source pipelines like AniMaker (62.2) and MovieAgent (65.3), even before iterative refinement. Scaling to $T=4$ via Random Search (75.7), akin to Best-of-N~\citep{snell2024scaling}, yields noticeable improvements. Crucially, replacing random sampling with our MAB formulation and warm-up (\moniker, $T=4$) drives the average to 81.4, demonstrating that our ultimate performance gains stem from a combination of superior reasoning, planning, and principled global optimization.

\begin{figure}[htbp]
    \centering
    \includegraphics[width=\linewidth]{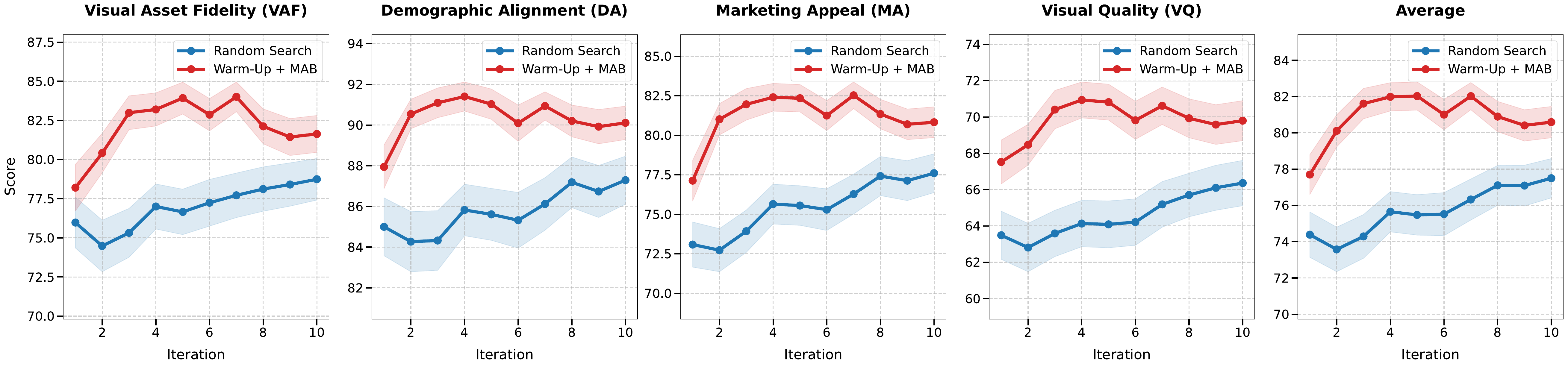}
    \caption{\textbf{MAB Optimization Efficiency.} Cumulative best performance over 10 iterations comparing \textsc{Co-Director}'s MAB optimization against an undirected Random Search.}
    \label{fig:mab_vs_random}
\end{figure}

As shown in Figure~\ref{fig:mab_vs_random}, comparing cumulative best performance reveals that \moniker’s \textit{Warm-Up + MAB} provides an efficiency advantage, achieving high-quality results in the first iteration where Random Search fails to ground the creative intent. While Random Search shows a steadier upward trajectory over long horizons, this reflects the variance of undirected sampling; it eventually "discovers" strong configurations by chance, whereas the MAB rapidly isolates an optimal plateau. This suggests the system effectively converges on a strategically sound local optimum early, avoiding the prohibitive computational costs of the exhaustive search required for Random Search to reach comparable parity.

\subsection{MLLM Metric Validation via Human Evaluation}
\label{sec:human_validation}
To validate the MLLM-as-a-Judge metrics, we conducted a human study on a 50-scenario subset. Each video was evaluated by 5 independent raters along the four dimensions (VAF, DA, MA, VQ) on a 5-point Likert scale. 

Table~\ref{tab:correlation} reports \textbf{Krippendorff’s Alpha ($\alpha$)}, \textbf{Cohen's Kappa ($\kappa$)}, \textbf{Pearson ($r$)}, \textbf{Spearman ($\rho$)}, and \textbf{Mean Absolute Error (MAE)}, establishing human agreement as the theoretical upper bound. The MLLM demonstrates strong alignment, approaching this ceiling in complex dimensions like DA and MA. Visual Quality (VQ) exhibits lower correlation due to differing evaluation mechanics: humans viewing real-time playback often find transient diffusion artifacts less visually disruptive. The MLLM, conversely, processes frames analytically, strictly penalizing brief structural or temporal inconsistencies. Thus, while serving as a highly aligned proxy for narrative evaluation, the MLLM's VQ score acts as a strict physical quality gate rather than a direct mirror of subjective human perception.

\begin{table}[htbp]
\centering
\small 
\setlength{\tabcolsep}{4pt} 
\caption{\textbf{Human agreement and correlation statistics across 50 scenarios.} Human-human agreement serves as the theoretical ceiling for MLLM-human alignment.}
\label{tab:correlation}
\begin{tabular}{l ccccc ccccc}
\toprule
\multirow{2}{*}{\textbf{Dimension}} & \multicolumn{5}{c}{\textbf{Human-Human (Ceiling)}} & \multicolumn{5}{c}{\textbf{MLLM-Human}} \\
\cmidrule(lr){2-6} \cmidrule(lr){7-11}
& \textbf{$\alpha$} & \textbf{$\kappa$} & \textbf{$r$} & \textbf{$\rho$} & \textbf{MAE} $\downarrow$ & \textbf{$\alpha$} & \textbf{$\kappa$} & \textbf{$r$} & \textbf{$\rho$} & \textbf{MAE} $\downarrow$ \\
\midrule
VAF & 0.606 & 0.465 & 0.733 & 0.635 & 0.644 & 0.521 & 0.330 & 0.601 & 0.522 & 1.016 \\
DA  & 0.641 & 0.523 & 0.751 & 0.617 & 0.548 & 0.592 & 0.438 & 0.647 & 0.512 & 0.770 \\
MA  & 0.620 & 0.489 & 0.737 & 0.673 & 0.596 & 0.562 & 0.386 & 0.570 & 0.499 & 0.848 \\
VQ  & 0.578 & 0.463 & 0.705 & 0.675 & 0.642 & 0.469 & 0.346 & 0.345 & 0.317 & 0.945 \\
\bottomrule
\multicolumn{11}{l}{\scriptsize $\alpha$ (Krippendorff's Alpha): Calculated over the full rater pool. $\kappa$ (Cohen's Kappa) \& MAE: Average of all pairwise} \\
\multicolumn{11}{l}{\scriptsize individual comparisons.  $r$ (Pearson) \& $\rho$ (Spearman): Calculated against the Human MOS.}
\end{tabular}
\end{table}

\begin{table}[htbp]
\centering
\small
\caption{\textbf{Mean Opinion Scores (MOS).} Average of 5 human raters per video on a [1--5] scale.}
\label{tab:human_mos}
\begin{tabular}{lccccc}
\toprule
\textbf{Method} & \textbf{VAF} & \textbf{DA} & \textbf{MA} & \textbf{VQ} & \textbf{Avg.} \\
\midrule
AniMaker       & 3.34 & 3.71 & 2.72 & 2.50 & 3.07 \\
MovieAgent     & 3.64 & 4.06 & 2.85 & 2.31 & 3.22 \\
Wan 2.6        & 3.84 & 3.65 & 3.13 & 3.29 & 3.48 \\
Kling 3.0 Omni & 3.93 & 4.13 & 3.07 & 3.21 & 3.59 \\
Veo 3.1        & 3.90 & 4.20 & 3.40 & 3.32 & 3.71 \\
\midrule
\rowcolor{gray!10} \textbf{\moniker (Ours)} & \textbf{4.22} & \textbf{4.41} & \textbf{3.65} & \textbf{3.58} & \textbf{3.96} \\
\bottomrule
\end{tabular}
\end{table}

\paragraph{Perceptual Comparison (MOS)}
The Mean Opinion Scores (MOS) in Table~\ref{tab:human_mos} corroborate our findings through automatic evaluation. \moniker\ achieves the highest overall perceptual quality (3.96), outperforming the strongest monolithic baseline, Veo 3.1 (3.71), and significantly exceeding specialized agentic systems like AniMaker (3.07) and MovieAgent (3.22). Notably, \moniker\ establishes state-of-the-art performance across all four individual dimensions. This confirms that our framework's global optimization effectively raises the narrative and visual ceiling of the underlying generative backbones.

\subsection{Ablation Studies}

\paragraph{MAB Logic} We compare our MAB search logic \textbf{(1)} against two variants. Reverting to a cold start \textbf{(2)} degrades performance, as lacking strong initialization produces higher-variance trajectories during early exploration. Crucially, collapsing the multi-dimensional arm-specific reward into a single scalar \textbf{(3)}—using the holistic score meant for final video selection—reduces VAF by 5.0 points. This validates our factored reward design. By leveraging the MLLM evaluator to decouple the overall video score from arm-specific rewards, the MAB isolates the exact parameters driving success (see App.~\ref{app:video_verifier_rubrics}).

\paragraph{Self-Refinement Loops} Removing Storyline Refinement \textbf{(4)} primarily penalizes semantic metrics like MA and DA, validating its impact on narrative coherence. Removing Keyframe Refinement \textbf{(5)} causes a steep 9.8-point VAF drop, confirming its vital role in preserving product visual identity. Stripping all local refinement and global optimization yields a ``barebones'' single-pass baseline \textbf{(8)} (67.2 Avg.). Our complete architecture outperforms this linear performance floor by a relative 17.6\%.

\begin{table}[t]
\centering
\small
\caption{\textbf{Ablation results} ($T=4$). \textbf{S-Ref}: Storyline and \textbf{K-Ref}: Keyframe Refinement.}
\label{tab:master_ablation}
\begin{tabular}{lccc|ccccc}
\toprule
\# & \textbf{S-Ref.} & \textbf{K-Ref.} & \textbf{MAB Logic} & \textbf{VAF} & \textbf{DA} & \textbf{MA} & \textbf{VQ} & \textbf{Avg.} \\
\midrule
\rowcolor{gray!10} \textbf{(1)} & \checkmark & \checkmark & \textbf{Ours} & \textbf{79.8} & \textbf{88.6} & 80.0 & \textbf{67.5} & \textbf{79.0} \\
\midrule
\textbf{(2)} & \checkmark & \checkmark & Cold Start & 78.2 & 87.8 & 77.7 & 65.6 & 77.3 \\
\textbf{(3)} & \checkmark & \checkmark & Scalar Reward & 74.8 & 87.9 & 77.8 & 65.3 & 76.4 \\
\midrule
\textbf{(4)} & $\times$ & \checkmark & Ours & 74.0 & 86.2 & 73.7 & 62.0 & 74.0 \\
\textbf{(5)} & \checkmark & $\times$ & Ours & 70.0 & 82.5 & 71.5 & 62.0 & 71.5 \\
\textbf{(6)} & $\times$ & $\times$ & Ours & 68.1 & 78.6 & 69.0 & 62.0 & 69.4 \\
\midrule
\textbf{(7)} & \checkmark & \checkmark & None & 66.2 & 75.3 & 67.6 & 63.6 & 68.2 \\
\textbf{(8)} & $\times$ & $\times$ & None & 65.4 & 74.8 & 66.1 & 62.5 & 67.2 \\
\bottomrule
\end{tabular}
\end{table}

\subsection{Qualitative Results and Discussion}
To complement our quantitative analysis, Appendix~\ref{app:qualitative_gallery} provides qualitative results showcasing generation artifacts across \benchmark scenarios. These examples demonstrate \moniker's ability to interpret abstract ideas into cohesive visual narratives with precise scene instructions. Our observations highlight three primary strengths: (1) \textbf{Strategic Adaptability}: evoking expected atmospheres for stereotypical prompts while maintaining consistency in unconventional contexts; (2) \textbf{Identity Preservation}: achieving character and product consistency across dynamic poses, varied angles, and dramatic environmental transitions; and (3) \textbf{Logical Robustness}: local self-refinement correcting LLM hallucinations like misidentified product categories (App.~\ref{app:self_refine}). Furthermore, evaluation on ViStoryBench-Lite~\citep{zhuang2025vistorybench} (App.~\ref{app:vistorybench_eval}) confirms \moniker's core consistency and creative steering generalize to non-advertising cinematic narratives, outperforming specialized baselines in style and character fidelity. We encourage readers to consult the appendix and supplementary videos for a complete trace of pipeline reasoning and rendered outputs.

\section{Conclusion}
We present \moniker, a hierarchical multi-agent framework that formulates video storytelling as a global optimization problem. By navigating a creative latent space via MAB-driven steering, our system achieves high-fidelity narrative consistency that traditionally requires large-scale production teams. Furthermore, to evaluate this capability, we introduce \benchmark, a rigorous dataset designed to assess end-to-end generative workflows under strict targeting and visual constraints. Experimental results confirm that \moniker outperforms both monolithic models and existing agentic pipelines, offering a principled, highly efficient approach to autonomous visual storytelling that generalizes beyond advertising to broader cinematic narratives.

\bibliography{refs}
\bibliographystyle{abbrvnat}

\appendix

\section{Appendix Overview}

To assist in navigating the supplemental materials, this section provides a directory of the contents contained within this appendix. Each section covers a distinct component of our results, dataset, or methodology:

\begin{itemize}
    \item \textbf{Section~\ref{app:qualitative_gallery}: Qualitative Results on \benchmark} --- Visual galleries demonstrating \moniker's ability to preserve brand identity across diverse narrative arcs.
    \item \textbf{Section~\ref{app:vistorybench_eval}: Evaluation on General Video Storytelling} --- Performance metrics and qualitative comparisons on the ViStoryBench benchmark~\citep{zhuang2025vistorybench}.
    \item \textbf{Section~\ref{app:dataset}: \benchmark Details} --- Details on dataset construction and descriptive statistics.
    \item \textbf{Section~\ref{app:baseline}: Baseline Implementation Details} --- Standardized configurations, prompt templates, and foundation models used for all evaluated baselines.
    \item \textbf{Section~\ref{app:mab_efficiency}: MAB Optimization Efficiency} --- Analysis of sample efficiency and the convergence of our multi-armed bandit creative steering.
    \item \textbf{Section~\ref{app:creative_direction}: Creative Direction Examples} --- Walkthroughs of hierarchical parameterization and how aesthetic archetypes influence visual output.
    \item \textbf{Section~\ref{app:self_refine}: Local Optimization via Self-Refinement} --- Evidence of the agentic loop's ability to correct structural hallucinations and identity drift.
    \item \textbf{Section~\ref{app:storyline_verifier_rubrics}: Storyline Evaluation Prompt} --- The system prompt and logic used to evaluate narrative coherence in a storyline.
    \item \textbf{Section~\ref{app:keyframe_verifier_rubrics}: Keyframe Evaluation Prompt} --- The system prompt used to assess visual consistency and character preservation in keyframes.
    \item \textbf{Section~\ref{app:video_verifier_rubrics}: Final Video Evaluation Prompt} --- The system prompt and rubric for scoring end-to-end video quality and creative alignment.
    \item \textbf{Section~\ref{app:rubrics}: \benchmark Evaluation System Prompts} --- The complete set of automated scoring rubrics for the proposed benchmark.
\end{itemize}

\section{Qualitative Results on \benchmark}
\label{app:qualitative_gallery}

This section presents qualitative results that demonstrate \moniker's ability to translate six-point product prompts and reference visuals (brand logos and products) into cohesive and high-fidelity visual narratives. 

\begin{figure}[bp]
    \centering 
    \includegraphics[width=\columnwidth]{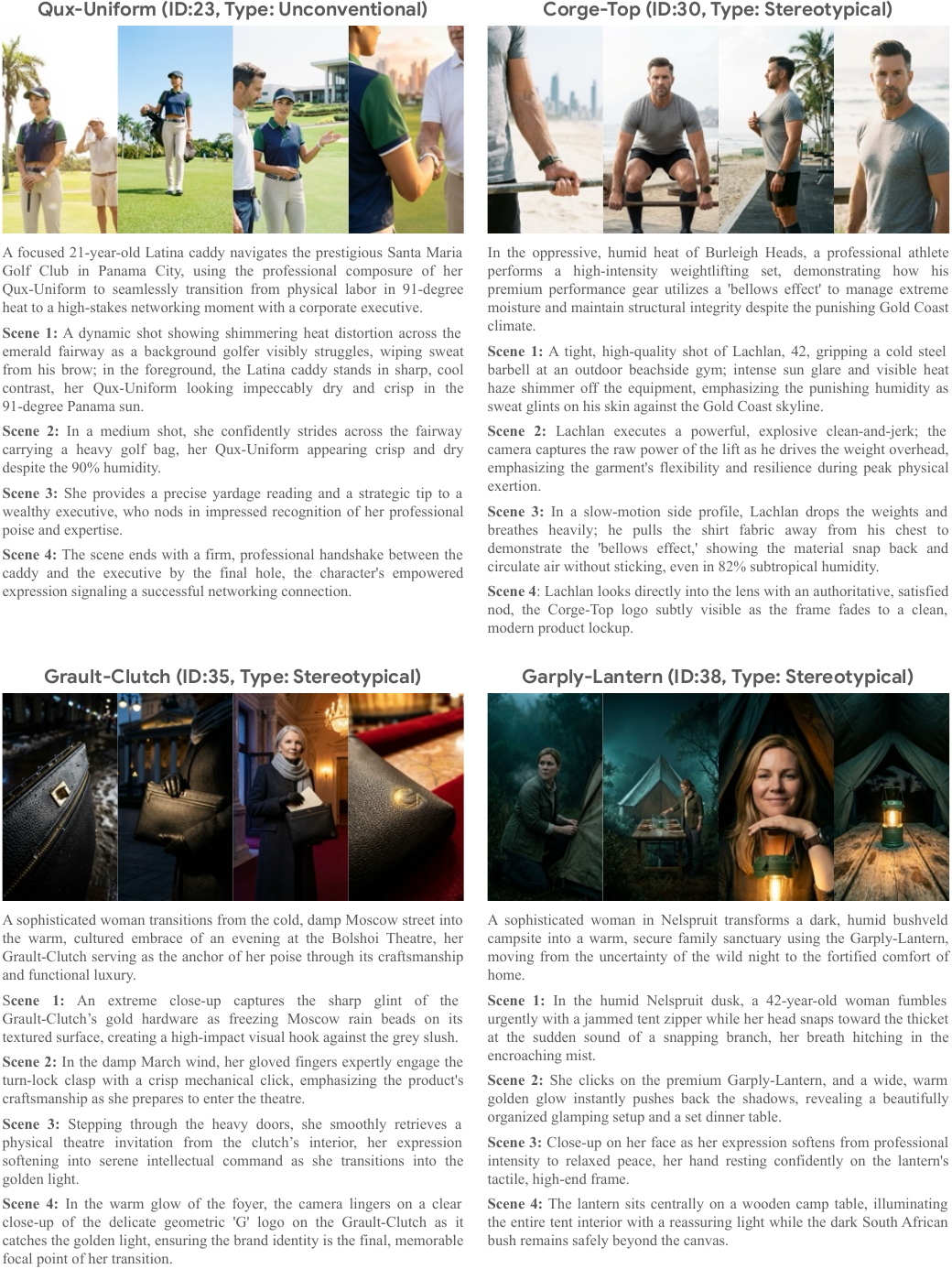}
    \caption{\textbf{Qualitative results on \benchmark.} \moniker translates complex product prompts into cohesive visual narratives. The framework maintains character and product identity across dynamic poses and varied angles (top) and preserves precise product details during dramatic environmental and lighting transitions (bottom). 
    }    \label{fig:gegnadbench_qualitative_1}
\end{figure}

Figure~\ref{fig:gegnadbench_qualitative_1} shows synthesized storylines alongside their corresponding keyframes across four distinct scenarios. These examples highlight the framework's ability to ground abstract marketing concepts---such as a functional 'bellows effect' or a psychological shift toward safety---into precise scene instructions with rich emotional arcs. Furthermore, they showcase the model's robust preservation of character and product identity across dynamic physical poses, varied camera angles, and dramatic environmental transitions.

\begin{figure}[bp]
    \centering 
    \includegraphics[width=0.98\columnwidth]{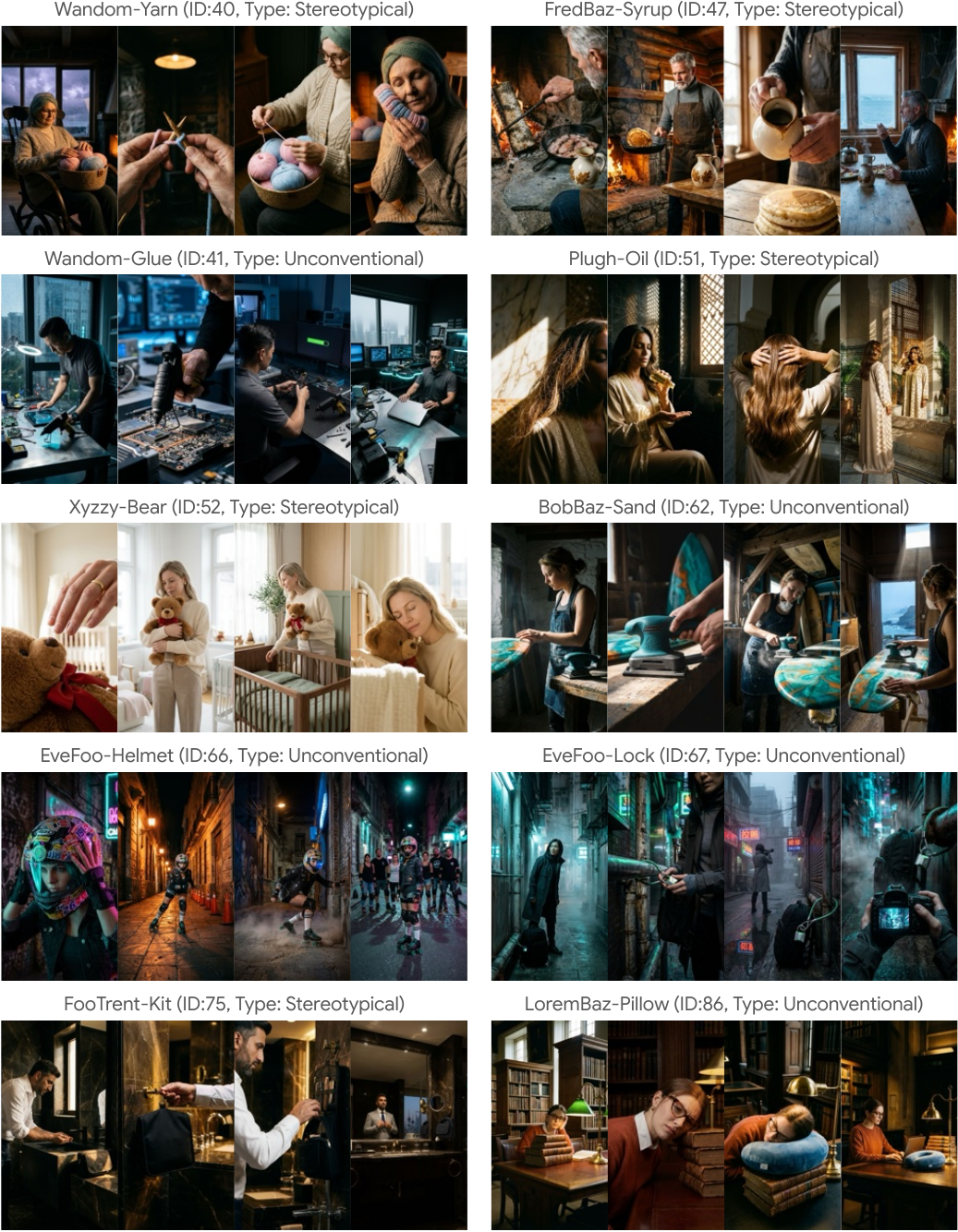}
    \caption{\textbf{Additional qualitative results on \benchmark.} \moniker effectively generates diverse visual narratives across both stereotypical and unconventional product scenarios. It successfully evokes expected atmospheres for stereotypical prompts, such as the rustic warmth of FredBaz-Syrup or the gentle nursery lighting of Xyzzy-Bear. The pipeline shows robustness to unconventional contexts, maintaining consistency in character/product identity and complex lighting (e.g., neon urban streets in EveFoo-Helmet) while executing precise object manipulation in unexpected environments like a library (LoremBaz-Pillow) or a high-tech workspace (Wandom-Glue).}
    \label{fig:gegnadbench_qualitative_2}
\end{figure}

To further illustrate the pipeline's versatility, Figure~\ref{fig:gegnadbench_qualitative_2} provides keyframe sequences from 10 additional scenarios, spanning both stereotypical and unconventional product contexts. \moniker consistently evokes the appropriate atmosphere for expected settings, such as rustic or softly lit environments, while maintaining structural and temporal consistency in  unconventional scenarios. This includes displaying the correct object manipulation in unexpected spaces (e.g., a high-tech workspace or a library) and handling complex lighting conditions like neon urban streets. 

\section{Evaluation on General Video Storytelling}
\label{app:vistorybench_eval}

In the main paper, we chose video advertising as our main demonstration scenario because it highlights a novel aspect of our framework: the ability to generate a full video narrative from a single, high-level idea, rather than needing a detailed, scene-by-scene script. Existing agentic systems, such as AniMaker~\citep{shi2025animaker} and MovieAgent~\citep{wu2025automated}, require structured, scene-by-scene scripts provided a priori. In contrast, our framework synthesizes full cohesive narratives from a brief tagline (e.g., our six-point prompt). While MAViS~\citep{wang2026mavis} attempts a similar idea-to-video pipeline, their evaluation relies on a closed set of 10 custom prompts, precluding reproducible comparison. This underscores the critical importance of our proposed \benchmark. Existing visual narrative datasets, such as MovieBench~\citep{wu2025moviebench} and ViStoryBench~\citep{zhuang2025vistorybench}, evaluate isolated generative steps (e.g., text-to-image or image-to-video synthesis) rather than the complete video storytelling pipeline. \benchmark\ fills this void by providing a rigorous, reproducible standard for evaluating the entire trajectory of video storytelling—from abstract conceptualization to final rendered output.

Despite the architectural mismatch between our end-to-end framework and traditional staged benchmarks, we evaluate \moniker\ on ViStoryBench-Lite~\citep{zhuang2025vistorybench} to demonstrate its generality across domains. Because ViStoryBench evaluates storyboard generation---providing strict shot-by-shot scripts and character reference images as inputs, and expecting static image outputs rather than video---we ablate our pipeline accordingly. Specifically, we bypass our script generation and video synthesis modules, isolating and deploying only the Keyframe Agent, the local self-refinement loop, and the global MAB-steered optimization. This allows us to prove that \moniker's core consistency and creative direction mechanisms remain highly effective even when constrained to rigid, pre-defined narrative structures. To ensure a rigorous evaluation, we deliberately select agentic video generation pipelines from the official ViStoryBench leaderboard as our baselines, as their modular architectures provide the most direct and equitable comparison to our own.

\begin{table}[bp]
\centering
\scriptsize 
\caption{\textbf{Evaluation on \textsc{ViStoryBench-Lite}}.} 
\setlength{\tabcolsep}{3.5pt} 
\begin{tabular}{l cc cc cc c ccccc}
\toprule
\multirow{2}{*}{\textbf{Method}} & \multicolumn{2}{c}{\makecell{\textbf{Style} \\ \textbf{Consistency}}} & \multicolumn{2}{c}{\makecell{\textbf{Character} \\ \textbf{Consistency}}} & \multicolumn{2}{c}{\makecell{\textbf{Quality \&} \\ \textbf{Diversity}}} & \multirow{2}{*}{\textbf{OCCM}} & \multicolumn{5}{c}{\textbf{Prompt Alignment}} \\
\cmidrule(lr){2-3} \cmidrule(lr){4-5} \cmidrule(lr){6-7} \cmidrule(lr){9-13}
& \textbf{Cross} & \textbf{Self} & \textbf{Cross} & \textbf{Self} & \textbf{Aesthetics} & \textbf{Inception} & & \textbf{Scene} & \textbf{Camera} & \textbf{GCA} & \textbf{LCA} & \textbf{Avg} \\
\midrule
MMStoryAgent & 0.261 & 0.661 & 0.385 & 0.598 & \textbf{5.910} & 8.090 & 58.236 & 2.965 & 2.452 & 1.643 & 1.500 & 2.012 \\
Vlogger & 0.299 & 0.497 & 0.373 & 0.519 & 4.242 & 8.827 & \textbf{79.145} & 1.749 & 3.143 & 2.429 & 2.350 & 2.404 \\
AniMaker & 0.305 & 0.558 & 0.423 & 0.593 & 5.603 & 9.942 & 72.034 & \textbf{3.622} & 2.932 & 3.388 & 2.850 & 3.128 \\
MovieAgent & 0.346 & 0.539 & 0.433 & 0.582 & 5.318 & \textbf{12.044} & 67.289 & 3.406 & 3.052 & \textbf{3.404} & 2.650 & 3.033\\
\bottomrule
\rowcolor{gray!10} \textbf{\moniker} & \textbf{0.499} & \textbf{0.743} & \textbf{0.499} & \textbf{0.593} & 5.286 & 10.900 & 66.215 & 3.412 & \textbf{3.288} & 2.999 & \textbf{2.966} & \textbf{3.166} \\
\bottomrule
\end{tabular}
\label{tab:main_evaluation}
\end{table}

As shown in Table~\ref{tab:main_evaluation}, \moniker\ achieves strong performance in visual consistency, securing the highest scores in Style (Cross: 0.499, Self: 0.743) and Character (Cross: 0.499, Self: 0.593) consistency metrics. Furthermore, it yields the highest average Prompt Alignment (3.166), demonstrating superior control over camera dynamics and local context compared to prior agentic baselines like AniMaker and MovieAgent. While \moniker\ trails in Quality \& Diversity (e.g., Inception Score: 10.900) and On-stage Character Count Matching (OCCM: 66.215), these metrics fundamentally conflict with cohesive storytelling. Inception Score favors high visual variance, naturally opposing our model's strict cross-scene consistency. Likewise, OCCM enforces a rigid ``closed-world'' character count, naturally penalizing \moniker\ for organically generating unprompted background actors (e.g., pedestrians in a ``bustling city'') that enhance cinematic realism.

\begin{figure}
    \centering 
    \includegraphics[width=\columnwidth]{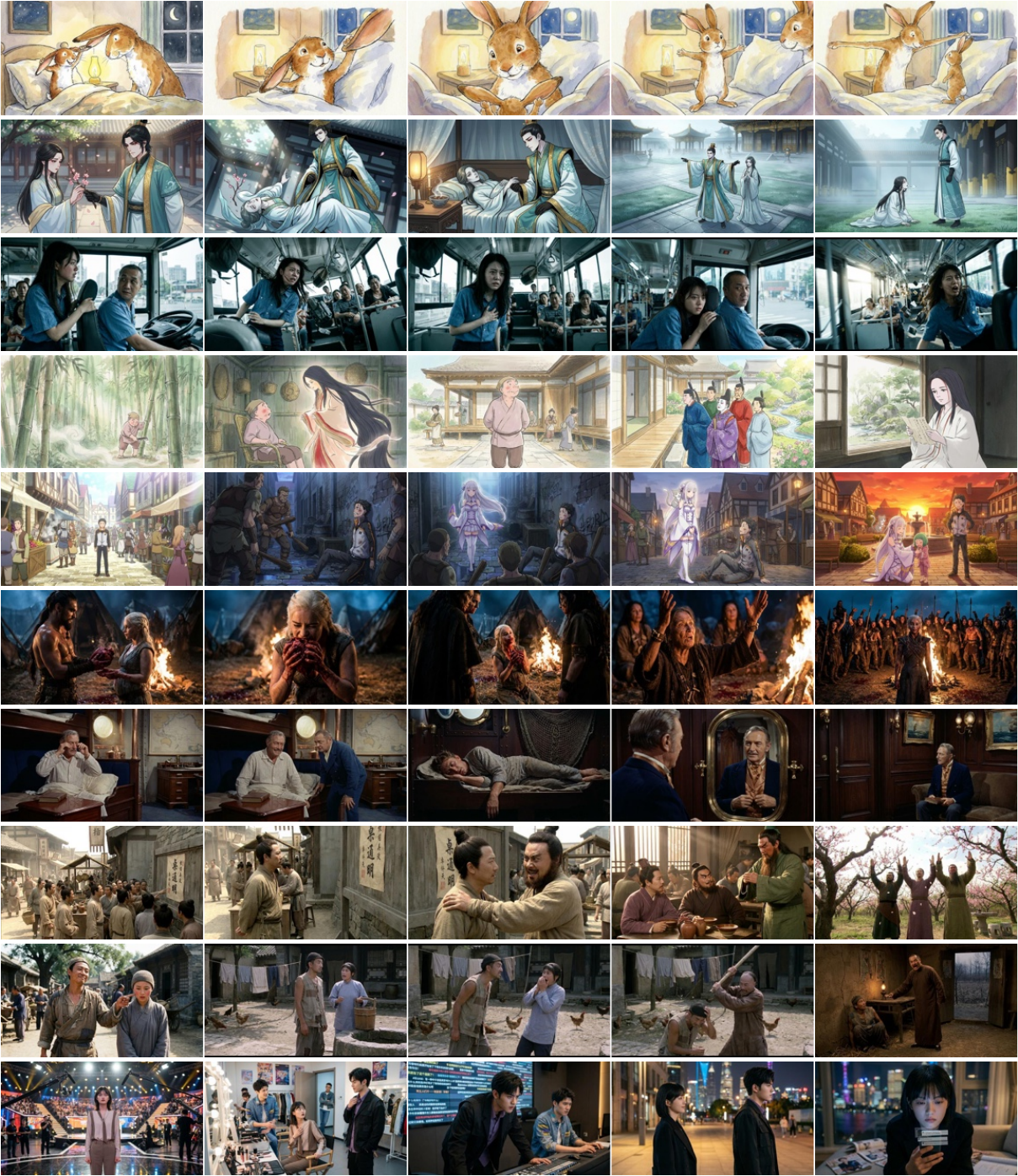}
    \caption{\textbf{Qualitative results on ViStoryBench.} Five consecutive frames are displayed for each generated story (01, 08, 27, 29, 32, 52, 55, 60, 64, 79). Across these diverse examples, \moniker\ maintains robust temporal consistency, successfully preserving both character identity and background spatial coherence throughout the sequences.}
    \label{fig:vistorybench_teaser}
\end{figure}

\begin{figure}
    \centering 
    \includegraphics[width=0.85\columnwidth]{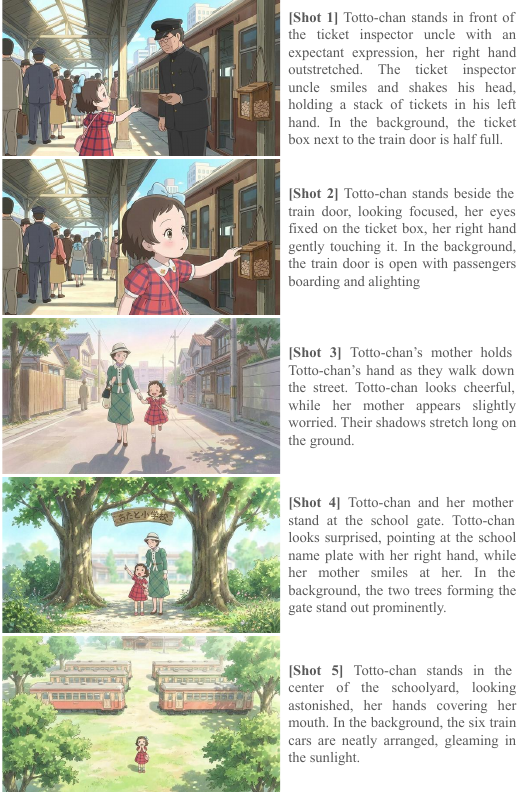}
    \caption{\textbf{Qualitative results on ViStoryBench (Story 28).} The generated sequence demonstrates strong prompt alignment and robust character consistency across frames, though minor anatomical artifacts (e.g., left-right hand confusion) occasionally occur.}
    \label{fig:vistorybench_28}
\end{figure}

These quantitative metrics are directly corroborated by qualitative observations. As illustrated in Figure~\ref{fig:vistorybench_teaser}, \moniker\ reliably generates temporally coherent frame sequences across diverse narratives, successfully preserving both character identity and background spatial structures over time. A closer examination of a specific sequence (Figure~\ref{fig:vistorybench_28}) further highlights the framework's strong prompt alignment and character preservation. While minor anatomical artifacts, such as occasional left-right hand confusion, can still manifest as an artifact of the underlying diffusion process, the framework's overall visual fidelity and narrative continuity remain exceptionally high.

Ultimately, these results illustrate that \moniker\ is not narrowly overfitted to marketing content, but is instead a highly capable engine for generalized visual storytelling. Crucially, video advertising should not be viewed as a restrictive niche; rather, it is a demanding subset of general storytelling characterized by rigid, testable constraints on brand fidelity, demographic alignment, and runtime. By mastering these rigorous constraints, \moniker\ demonstrates a foundational narrative competency that seamlessly transfers to broader cinematic and storytelling domains.

\section{\benchmark Details}
\label{app:dataset}

To rigorously evaluate video storytelling, we introduce \benchmark. This appendix details the dataset's construction process, comprehensive demographic and category statistics, our data split strategy, and qualitative examples.

\begin{figure}
    \centering 
    \includegraphics[width=0.98\columnwidth]{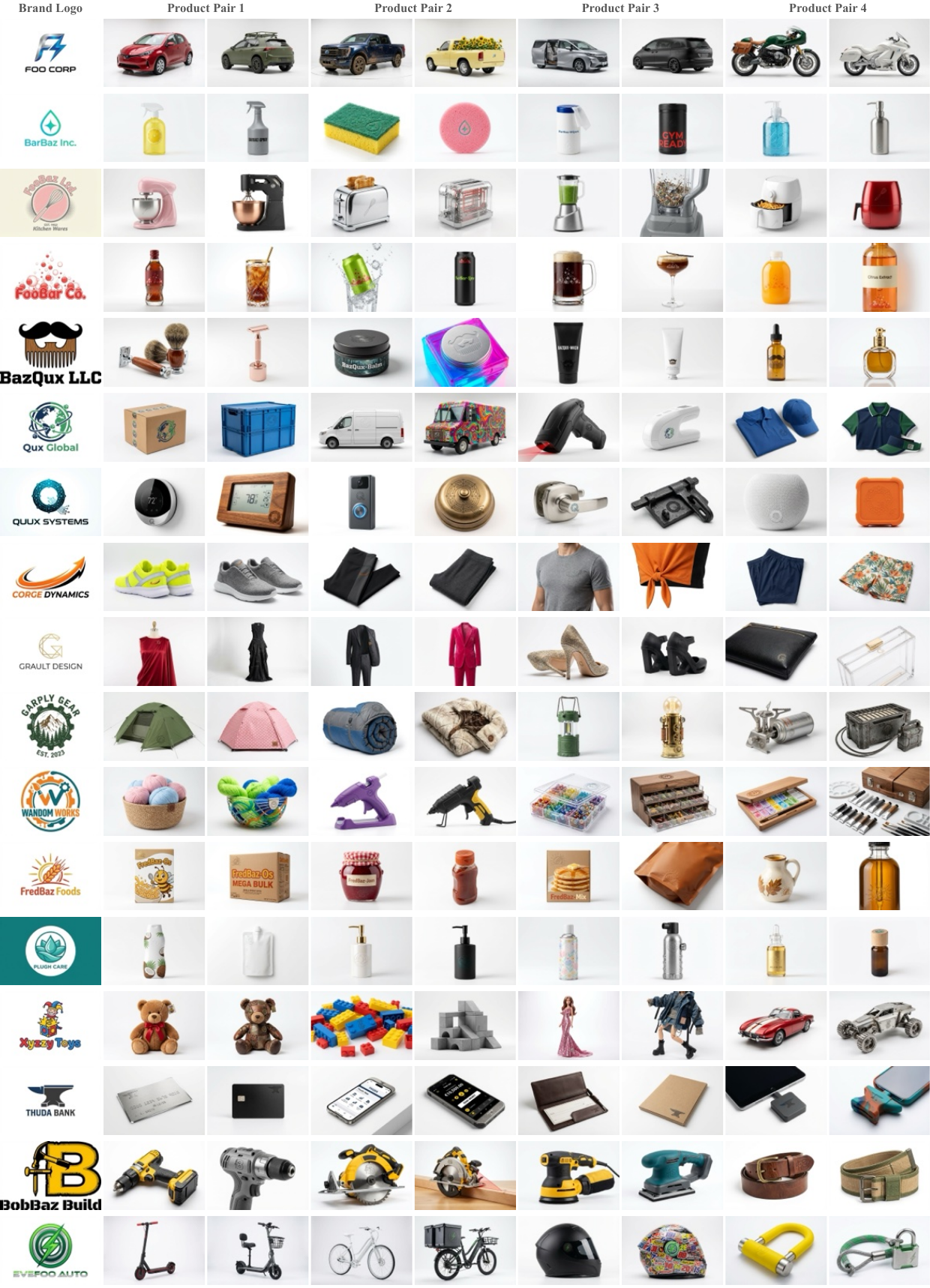}
    \caption{\textbf{\benchmark reference visuals.} The reference set comprises brand logos and product images, with each brand featuring four product pairs specifically curated to represent both stereotypical and unconventional target demographics.}
    \label{fig:genadbench_teaser}
\end{figure}

\subsection{Dataset Construction Pipeline}
Current video evaluation benchmarks predominantly rely on generic, short-horizon prompts (e.g., ``a dog running on grass'') and evaluate well-known entities, allowing models to rely heavily on parametric memory. \benchmark\ circumvents this by utilizing a human-in-the-loop pipeline to synthesize \textbf{200 fictitious products} across \textbf{50 fictitious brands}. 

We utilized \textsc{Gemini 3 Pro~\cite{gemini3p}} to generate the structural text metadata and \textsc{Nano Banana Pro~\citep{gemini3p_image}} to generate the visual reference assets. The construction followed a hierarchical schema:
\begin{enumerate}
    \item \textbf{Brand \& Product Synthesis:} The LLM generated 50 unique brands and assigned exactly 4 distinct products to each. Products were explicitly prompted to span a wide range of form factors, including small consumer goods, large industrial equipment, and intangible offerings (e.g., software and subscription services).
    \item \textbf{Paired Scenario Generation:} For each of the 200 products, the LLM generated two contrasting target demographics to test conceptual flexibility: a \textit{Stereotypical} scenario (aligned with traditional marketing tropes) and an \textit{Unconventional} scenario (subverting traditional expectations).
    \item \textbf{Visual Asset Generation:} The text-to-image model generated a distinct vector-style logo for each of the 50 brands. Subsequently, it generated product reference images conditioned on both the textual product description and the respective brand logo to ensure visual-semantic integration.
    \item \textbf{Manual Verification \& Quality Control:} To ensure high-fidelity inputs for the product reference generation stage, we first manually verified and corrected all brand logos. Following the generation of these product reference visuals, approximately 15\% were regenerated to address improper logo integration, logical inconsistencies, or unintended visual memorization (e.g., designs too closely resembling products existing in the real-world). This targeted refinement ensures the uniqueness and brand accuracy of the primary visual assets before they are propagated downstream.
\end{enumerate}

\subsection{Taxonomy and Distribution Statistics}
\benchmark\ contains 400 evaluation scenarios distributed across a diverse range of industries and global contexts. 

\begin{figure*}[t]
    \centering
    \includegraphics[width=\textwidth]{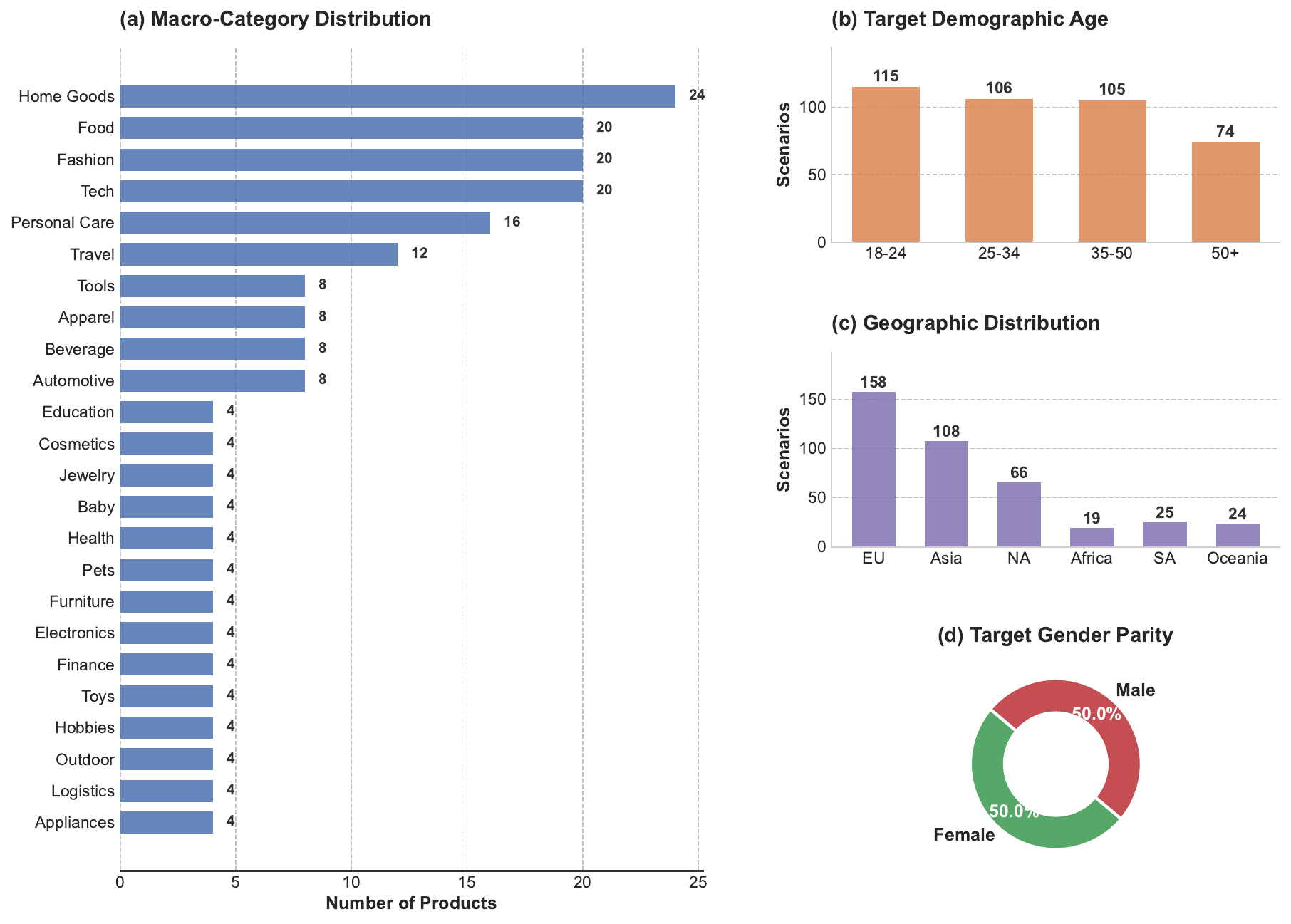}
    \caption{\textbf{Dataset Statistics for \benchmark.} The benchmark enforces rigorous categorical, demographic, and geographic diversity to prevent generative collapse. (a) The macro-category distribution spans a wide array of physical domains. (b) Target age demographics cover four distinct life stages. (c) Scenarios are situated across 183 unique global locations, grouped here into six continental macro-regions. (d) Target gender distributions are perfectly balanced at the product level to neutralize latent generation biases.}
    \label{fig:dataset_stats}
\end{figure*}

\paragraph{Macro-Category Distribution}
The dataset challenges models to render various physical properties, including reflective metals (Jewelry), transparent liquids (Perfumes), and complex dynamics (Drones, Apparel). As illustrated in Figure~\ref{fig:dataset_stats}a, the 200 products span 24 distinct macro-categories. High-frequency categories such as Home Goods (24), Food (20), Fashion (20), and Tech (20),  test foundational object generation, while specialized domains like Automotive (8), Logistics (4), and Baby (4) probe the model's ability to synthesize niche, domain-specific contexts without hallucination.

\paragraph{Demographic and Geographic Parity}
To ensure the multi-armed bandit optimization does not merely exploit ingrained training biases, \benchmark\ explicitly enforces balanced demographic and geographic distributions. As shown in Figure~\ref{fig:dataset_stats}d, the dataset maintains exact gender parity: 50\% (200) of the target personas are Male, and 50\% (200) are Female. Crucially, because paired scenarios invert the demographic targets for the exact same product, gender bias is neutralized at the product level. Target age brackets (Figure~\ref{fig:dataset_stats}b) uniformly represent four key life stages: Gen~Z (18--24), Millennials (25--34), Gen~X (35--50), and Seniors (50+). Finally, the framework demands robust cultural and environmental alignment beyond traditional Western-centric backdrops; the scenarios are grounded in 183 uniquely prompted global locations. While the dataset spans all six major continents, it deliberately concentrates on the two most diverse global consumer markets: Europe (158 scenarios) and Asia (108 scenarios). Within these clusters, geographic representation remains highly granular. The European scenarios span from primary Western hubs (e.g., France, UK, Germany) to Nordic and Eastern European locales, while the Asian distribution covers major East Asian centers (Japan, China, South Korea) alongside robust representation across Southeast Asia, India, and the Middle East.

\subsection{Dataset Split}
We partitioned the 400-scenario dataset into two equal 200-scenario splits. The \textbf{Hillclimbing Set} is reserved strictly iterative system optimization and is excluded from the evaluations presented in this work. All performance metrics are reported solely on the \textbf{Validation Set}, which is stratified into two sub-challenges to assess \moniker's generalization capabilities across 45 unique brands:
\begin{itemize}
    \item \textbf{In-Domain (80\% of Val / 160 Scenarios):} Contains 80 products from 40 brands that share macro-categories with the Hillclimbing set. This evaluates a system's ability to extrapolate optimized strategies to entirely novel products and brands within familiar industries.
    \item \textbf{Out-of-Domain (20\% of Val / 40 Scenarios):} Contains 20 products from 5 brands drawn from distinct industries (e.g., Pets, Jewelry) that are absent from the Hillclimbing set. This tests a system's broader generalization capabilities when handling product categories for which no category-specific optimization has occurred.
\end{itemize}

\section{Baseline Implementation Details}
\label{app:baseline}

\paragraph{Monolithic and Commercial Platforms} To evaluate the monolithic foundation models and specialized commercial platforms---\textbf{LTX-2.3}~\citep{hacohen2024ltx}, \textbf{Kling 3.0 Omni}~\citep{kling2026kling3omni}, \textbf{Veo 3.1}~\citep{veo3.1}, \textbf{Wan 2.6}~\citep{wan2025wan}, 
\textbf{Creatify}~\citep{creatify2026}, and \textbf{HeyGen}~\citep{heygen2026}---we employ a standardized zero-shot prompting strategy. Each system is queried using the template: \texttt{``Generate a video advertisement. [SIX-POINT PROMPT]. Reserve the last 1 second for an ending scene highlighting the provided brand logo image.''} This ensures a uniform evaluation setting that  tests each baseline's ability to interpret the core product brief and integrate branding constraints. All the baseline methods natively generate audio given text prompt.

\paragraph{Multi-Agent Frameworks} We compare against two structurally analogous agentic pipelines, \textbf{AniMaker}~\citep{shi2025animaker} and \textbf{MovieAgent}~\citep{wu2025automated}, to isolate the architectural efficacy of our multi-agent system from the underlying generative backbones. Because these baselines require structured, multi-shot scripts rather than abstract product briefs, we introduced a Script Generator module. This module leverages \textsc{Gemini 3 Pro}~\citep{gemini3p} to decompose the initial prompt into a structured JSON script. The resulting script explicitly dictates the narrative arc, scene-specific visual prompts, voice-over text, and precise shot durations, enforced by a strict 12-second total video length constraint. To maintain a controlled comparison against our framework, these baselines are instantiated using identical foundation models: Veo 3.1~\citep{veo3.1} for video generation and Gemini 2.5 Pro TTS~\citep{gemini2.5p_tts} for voiceovers

\section{Extended Discussion on MAB Optimization Efficiency}
\label{app:mab_efficiency}

While standard multi-armed bandit (MAB) formulations often analyze asymptotic convergence over thousands of iterations, video generation imposes extreme computational and financial constraints. An exhaustive grid search of our defined creative action space ($\theta_{cs} \times \theta_{nm} \times \theta_{aa}$) would require 36 discrete, end-to-end video generation pipelines per prompt. While this brute-force approach guarantees finding the global optimum, it renders the system entirely unscalable and defeats the purpose of an accessible agentic framework. 

Therefore, the goal of our MAB formulation is not infinite-horizon exploration, but rather sample efficiency under resource constraints. We achieve this through two mechanisms:
\begin{enumerate}
    \item \textbf{Factored Rewards:} By utilizing a factored reward signal $R = (r_{cs}, r_{nm}, r_{aa})$, a single execution step $t$ simultaneously updates the expected values across all three independent creative axes. The system is not searching 36 distinct paths, but rather evaluating 10 independent parameters.
    \item \textbf{LLM-Driven Warm Start:} We initialize the expected values using an LLM, safely biasing early exploration toward theoretically sound configurations and pruning the vast majority of sub-optimal combinations before iteration even begins.
\end{enumerate}

Consequently, as demonstrated in our empirical results (Fig.~\ref{fig:mab_vs_random}), an iteration budget of $T=4$ is sufficient for the MAB to rapidly converge on a highly effective creative configuration. This approach clearly outperforms Random Search, capturing the benefits of a global search without incurring the computational overhead of exhaustive sampling.

\section{Creative Direction Examples}
\label{app:creative_direction}

\subsection{Horizontal Directing through Hierarchical Parameterization}
\moniker\ achieves global coherence through \textit{horizontal directing}—the process of synthesizing disparate creative parameters into a unified, high-level vision. Rather than passing raw categorical arms as static strings, the Orchestrator Agent ($\Phi_{orch}$) generates a top-down creative direction that is decomposed into agent-specific technical directives. These directives are then routed to their respective sub-agents via dynamic prompt injection; specifically, the synthesized creative direction is injected into the system prompt of each agent. This ensures that the Storyline, Keyframe, and Video instructions are orthogonally aligned toward a single strategic goal, preventing semantic drift by grounding each agent’s execution in a centralized creative intent.

Table~\ref{tab:parameter_synthesis} illustrates an example. In Iteration 1, the bandit selects an \textit{Informational} strategy ($\theta_{cs}$) delivered via a \textit{Vignette} narrative mode ($\theta_{nm}$) and \textit{Clarity/Energy} ($\theta_{aa}$) aesthetics. This results in directives that prioritize "functional utility" and "vibrant suburban transit," utilizing high-temporal motion and fast cuts to emphasize the efficiency of the vehicle's sliding doors and safety frame within an Osaka carpool context.

In contrast, in Iteration 2, the bandit pivots to a \textit{Transformational} strategy ($\theta_{cs}$) using an \textit{Analytical} mode ($\theta_{nm}$) and \textit{Cinematic Premium} ($\theta_{aa}$). The instructions are immediately overridden to focus on a "logical manifesto" of access and protection, delivered through a "Premium Noir" aesthetic. This configuration utilizes Chiaroscuro lighting and low-temporal, reflective cinematography to connect the vehicle’s technical specifications to the user’s social identity and psychological state of "unshakable composure." This horizontal consistency across different narrative and sensory modalities demonstrates that \moniker is capable of high-level creative reasoning, ensuring that the "what," "how," and "feel" of the advertisement remain in strict strategic alignment.

\begin{table}[tp]
\centering
\caption{\textbf{MAB Parameter Synthesis into Creative Directions.} This table illustrates how specific combinations of \textbf{Creative Strategy} ($\theta_{cs}$), \textbf{Narrative Mode} ($\theta_{nm}$), and \textbf{Aesthetic Archetype} ($\theta_{aa}$) yield differentiated directives across the Story, Keyframe, and Video agents. Example: \textit{Foo-Family by Foo Corp} (ID: 2, Type: Stereotypical).}
\label{tab:parameter_synthesis}
\scriptsize
\begin{tabularx}{\textwidth}{X|X|X}
\toprule
\textbf{Pre-Production Agent (Storyline)} & \textbf{Keyframe Agent} & \textbf{Video Agent} \\
\midrule
\multicolumn{3}{c}{\cellcolor{gray!15}\textbf{Iteration 1} --- $\theta_{cs}$: Informational — $\theta_{nm}$: Vignette — $\theta_{aa}$: Clarity/Energy} \\
\midrule
Develop a script based on the Informational Creative Strategy (Laskey et al., 1989), focusing on the functional utility of the Foo-Family vehicle. The narrative must follow the 'Slice of Life' mode (Escalas, 2004), presenting a series of atmospheric vignettes of adult women (aged 35-50) in Osaka, Japan, engaged in a carpool routine. Each scene must highlight a specific product attribute: the ease of the dual sliding doors for loading equipment, the spacious multi-passenger seating for adult commuters, and the reinforced safety frame. Avoid a traditional plot arc; instead, create a consistent 'vibe' of efficient, safe suburban transit. Use text overlays to provide factual evidence of safety ratings and door clearance measurements to appeal to the logical advantages of the vehicle. & 
Generate high-key, low-contrast imagery consistent with Zettl’s (1973) Media Aesthetics to suggest transparency and vitality. The visual style should be clean, modern, and bright, featuring the Foo-Family vehicle in suburban Osaka settings. Focus on the textures of the vehicle's interior and the mechanical precision of the sliding doors. Characters should be Japanese women aged 35-50, dressed in contemporary professional-casual attire, interacting with the car in a bright, airy environment. Use a color palette of whites, soft blues, and metallic silver to emphasize safety and cleanliness. Ensure the Foo Corp logo is clearly visible on the vehicle. & 
Execute a high-temporal motion edit characterized by fast cuts and high-arousal transitions (Zettl, 1973). The cinematography should utilize dynamic camera movements, including quick tracking shots of the sliding doors in motion, rapid dolly-ins to safety sensors, and whip-pans between adult passengers to create a sense of energy and efficiency. The frame rate should be standard, but the editing rhythm must be upbeat and synchronized with a high-tempo, optimistic audio track. Use bright, even lighting throughout to maintain a high-key aesthetic. The camera should emphasize the vehicle's accessibility and safety features through close-ups of functional components, maintaining a fast-paced, vibrant flow that reflects the vitality of the Osaka carpool lifestyle. \\
\midrule
\multicolumn{3}{c}{\cellcolor{gray!15}\textbf{Iteration 2} --- $\theta_{cs}$: Transformational — $\theta_{nm}$: Analytical — $\theta_{aa}$: Cinematic Premium} \\
\midrule
Develop a script based on an argument-based narrative mode (Escalas, 2004) that delivers a transformational creative strategy (Laskey et al., 1989). The script must function as a logical manifesto rather than a story, using direct address to the viewer. Focus on three logical pillars: 'The Logic of Access' (sliding doors), 'The Logic of Protection' (safety features), and 'The Logic of Community' (multi-passenger capacity). Each point must be framed to evoke a psychological state of 'unshakable composure' for the Osaka-based female lead (age 35-50). The tone is premium, intellectual, and emotionally resonant, connecting the vehicle's specs to the user's social identity as a reliable pillar of her suburban community. &
Generate high-fidelity scenes utilizing Chiaroscuro lighting (Zettl, 1973). Visuals must feature high contrast with deep, dramatic shadows and sharp, intentional highlights that define the Foo-Family vehicle's silhouette and sliding door mechanism. The setting is a refined suburban Osaka environment at dusk, emphasizing a 'Premium Noir' aesthetic. Surfaces should show realistic reflections of city lights. Characters are adult females (35-50) with elegant, stoic expressions, dressed in high-end professional or suburban attire. Ensure the Foo Corp logo and vehicle branding are visible but integrated into the high-contrast environment.  &
Execute cinematography with low-temporal motion (Zettl, 1973). Use slow, deliberate dolly-in and tracking shots that glide alongside the vehicle at a reflective pace. The motion of the sliding doors must be captured in smooth, weighted slow-motion to emphasize engineering precision. Avoid rapid cuts or handheld jitter; every frame must feel stable and composed. The camera should focus on the textures of the vehicle and the calm, confident faces of the adult passengers. Audio must be a cinematic, ambient orchestral score—heavy on strings and low-frequency resonance—to reinforce the high-end, transformational feel. The pacing must be rhythmic and analytical, matching the logical progression of the argument-based script.\\
\midrule
\multicolumn{3}{c}{\cellcolor{gray!15}\textbf{Iteration 3} --- $\theta_{cs}$: Comparative — $\theta_{nm}$: Narrative Drama — $\theta_{aa}$: Minimalist Focus} \\
\midrule
Develop a three-act narrative script centered on a 40-year-old female driver in a bright, suburban Osaka neighborhood. Act 1 (Conflict): The protagonist struggles with the 'urban norm'—a standard sedan with hinged doors parked in a narrow space, making it impossible to load large school event supplies and for her adult friends to enter. Act 2 (Resolution): Introduce the Foo-Family vehicle. Highlight the unique value proposition of the motorized sliding doors that glide open effortlessly in the same tight space. Act 3 (Transformation): The group of adult women sits comfortably inside the spacious cabin, sharing a moment of relief and safety. The script must emphasize the contrast between the cramped standard and the Foo-Family's accessibility.&
Apply a Zettl-inspired 'Applied Media Aesthetics' style. Visuals must be high-key with bright, clean, and airy backgrounds that evoke a pristine Osaka suburban morning. Focus on extreme detail and tactile textures: the brushed metal of the Foo-Family logo, the smooth track of the sliding door, and the premium fabric of the interior seating. Use a minimalist color palette with soft whites, silvers, and light blues. Every frame should look hyper-real and clinical yet inviting. All human subjects must be adult females aged 35-50, dressed in sophisticated, contemporary Japanese fashion. &
Execute a cinematography plan focused on micro-movements and static precision. Use slow-motion macro shots to capture the mechanical elegance of the sliding door opening. Camera movement should be limited to subtle, buttery-smooth dollies or slow pans that emphasize the vehicle's length and safety features. Avoid handheld or shaky movements. The visual rhythm should be calm and deliberate, allowing the viewer to focus on the 'ASMR' quality of the car's operation—the click of a seatbelt, the whisper-quiet slide of the door, and the tactile touch of the dashboard. Ensure the lighting remains consistently high-key throughout all transitions \\
\bottomrule
\end{tabularx}
\end{table}

\subsection{Visual Realization via Aesthetic Archetypes}
\begin{figure}[tp]
\centering
    \begin{subfigure}{.9\textwidth}
        \centering
        \includegraphics[width=\linewidth]{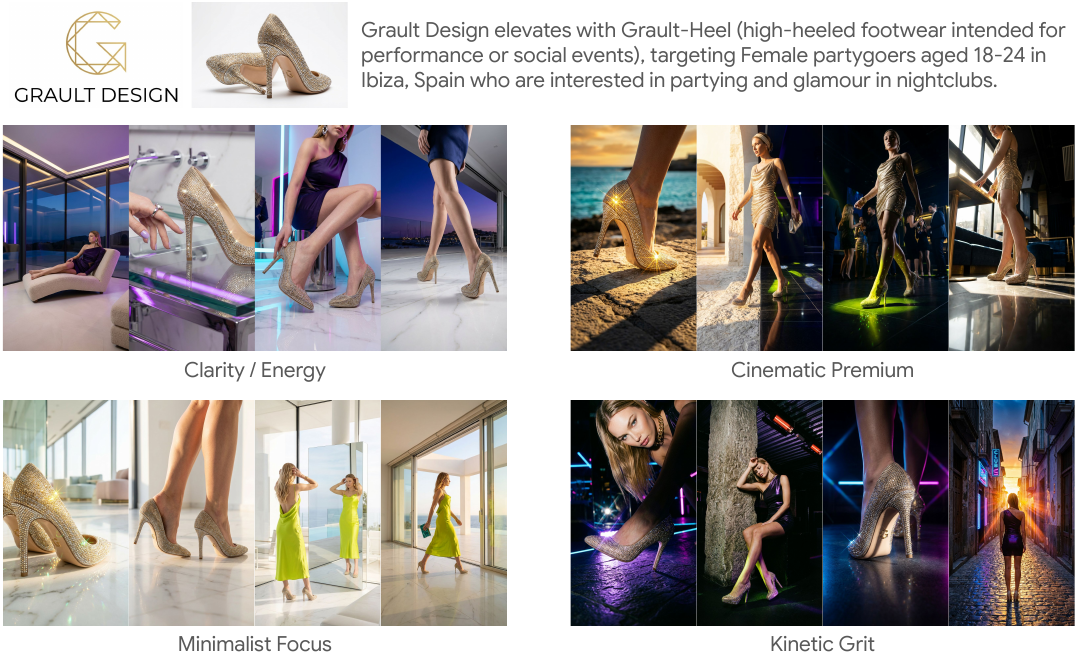}
        \vspace{1pt}
    \end{subfigure}
    \begin{subfigure}{.9\textwidth}
        \centering
        \includegraphics[width=\linewidth]{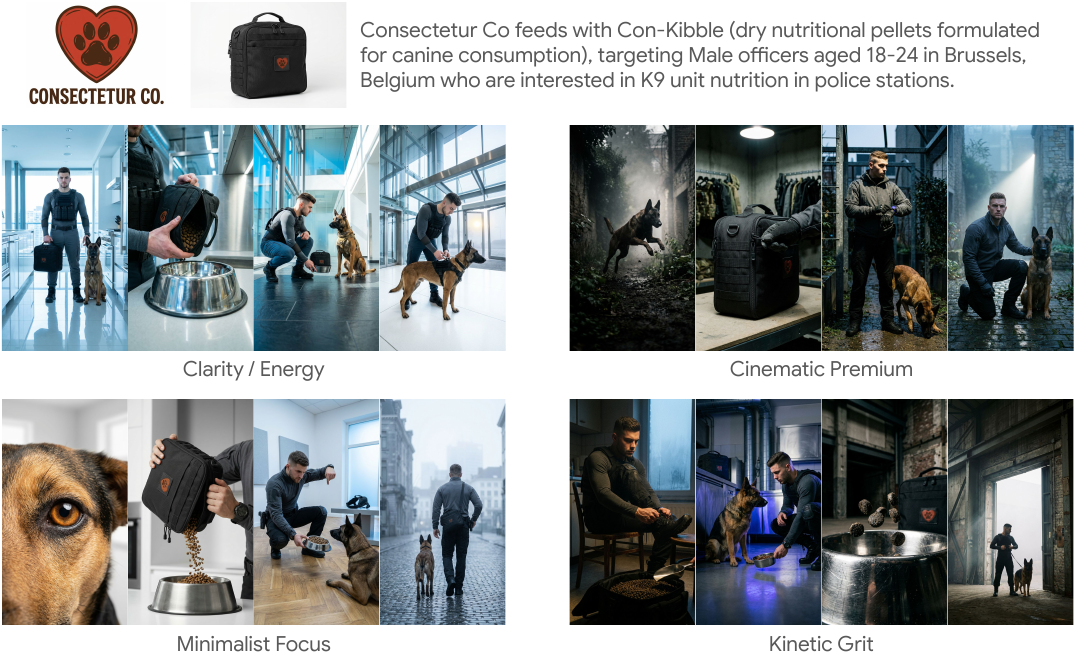}
    \end{subfigure}
    \caption{\textbf{Visual Guide to Aesthetic Archetypes ($\theta_{aa}$):} This comparison shows how the same product is "re-shot" by the system to match different moods.
\textbf{(a) Clarity/Energy ("The Pop Look"):} Everything is bright and easy to see. It uses "high-key" lighting with no dark shadows to make the product look clean, vibrant, and exciting.
\textbf{(b) Cinematic Premium ("The Movie Look"):} This uses "Chiaroscuro" (dramatic light and deep shadows). It makes the product look expensive and legendary, like a hero in a high-end film.
\textbf{(c) Minimalist Focus ("The Gallery Look"):} The background is deleted or made clean. The camera zooms in to show off tiny textures (like crystals or kibble pellets) in a calm, "museum-like" setting.
\textbf{(d) Kinetic Grit ("The Raw Look"):} This feels like a real-world documentary. The lighting is dark and "moody," and the camera feels like it's being held by a person (handheld) or a drone, creating a sense of raw intensity. Examples: \textit{Grault-Heel by Grault Design} (ID: 34, Type: Stereotypical) and \textit{Con-Kibble by Consectetur Co.} (ID: 96, Type: Unconventional). } 
\label{fig:appendix_aesthetic_archetypes}
\end{figure}

To evaluate the qualitative impact of our hierarchical parameterization, we analyze the visual output of two scenarios across four distinct \textit{Aesthetic Archetypes} ($\theta_{aa}$). As shown in Figure~\ref{fig:appendix_aesthetic_archetypes}, the MAB-selected configuration acts as a "Digital Director," re-shooting the same product prompt to meet specific strategic intents:

\begin{itemize}[leftmargin=15pt, itemsep=5pt, parsep=0pt, topsep=1pt]
    \item \textbf{Clarity/Energy (the "Pop" Look):} This maximizes immediate product recognition. By flooding the scene with high-key, low-contrast illumination, the system eliminates shadows and emphasizes vibrancy. In the \textit{Grault-Heel} example, this manifests as clean, energetic neons; in the \textit{Con-Kibble} case, it produces a clinical, trustworthy brightness.
    
    \item \textbf{Cinematic Premium (the "Movie" Look):} This targets perceived prestige via \textit{chiaroscuro} (high-contrast) lighting. By sculpting depth through dramatic light and shadow, the product is elevated to a "hero" status. Note the sophisticated golden-hour rim lighting on the footwear and the moody, atmospheric shadows in the K9 training station.
    
    \item \textbf{Minimalist Focus (the "Gallery" Look):} Here, the system prioritizes material essentialism. By stripping away environmental noise and utilizing macro-framing against pure backgrounds, the viewer is forced to engage with the physical "truth" of the object—such as the intricate crystal weave of a heel or the nutritional texture of a kibble pellet.
    
    \item \textbf{Kinetic Grit (the "Raw" Look):} This emphasizes modern authenticity over polished perfection. Through low-key lighting and unstable handheld or FPV drone motion cues, the system captures a sense of "verité" intensity. This is visible in the gritty, urban shadows of the Ibiza nightlife and the tactical, unpolished urgency of the professional K9 unit.
\end{itemize}

Crucially, despite these shifts in aesthetic "vibe," the underlying reference visual assets remain consistent. The brand logos and structural features of the assets are preserved across every narrative arc, proving that our agentic steering does not compromise brand integrity.

\section{Local Optimization via Agentic Self-Refinement}
\label{app:self_refine}

Building on the foundations of feedback descent~\citep{yuksekgonul2024textgrad,lee2025feedback}, we implement an agentic self-refinement loop to ensure the logical and visual integrity of intermediate artifacts. While Section~\ref{sec:approach_refine} detailed the adaptation of this framework for our pipeline, this section provides qualitative evidence of its performance. Specifically, we demonstrate how the system identifies and corrects structural hallucinations in storylines and enforces character consistency across generated keyframes.

\subsection{Storyline Self-Refinement}
Following the multi-agent architecture described in Section~\ref{sec:mas_architecture}, the system first constructs a scene-by-scene storyline ($\mathcal{L}$). Because this script establishes the foundation for all subsequent visual assets, it is subjected to an automated refinement loop to ensure it strictly adheres to the product brief ($\mathcal{B}$). It is well-known that LLMs occasionally "hallucinate" product categories or misinterpret functional attributes. In the example below, the initial draft mistakenly identifies the \textsc{AlphaFoo} computer chassis as a motorized standing desk. The refinement agent identifies this category error and rewrites the scene to correctly feature the spherical hardware, preventing a logic gap that would otherwise carry through to the storyboard and video generation phases.

\begin{tcolorbox}[colback=blue!5, colframe=blue!30, title={\textbf{Six-point Prompt} (ID: 114, Type: Unconventional)}, fonttitle=\bfseries]
    \small AlphaFoo Corp builds AlphaFoo-Desk (stationary personal computer chassis housing processing components), targeting Female editors aged 35-50 in Burbank, California who are interested in rendering 4K video in video editing suites.
\end{tcolorbox}

\begin{tcolorbox}[colback=gray!5, colframe=gray!50, title=\textbf{Initial Storyline (before)}, fonttitle=\bfseries]
    \small A high-end Burbank editor transitions from the chaotic physical strain of a long 4K render session to a state of absolute 'flow' and ergonomic bliss at her AlphaFoo-Desk.
    
    \medskip
    
    \small \textbf{Scene 1:} A 42-year-old female editor in a sleek Burbank studio leans into her monitors, her face illuminated by the blue light of a complex 4K video timeline, showing visible signs of physical tension.
    
    \medskip
    
    \small \textbf{Scene 2:} The camera pans across the AlphaFoo-Desk as it smoothly adjusts to her perfect ergonomic height, her hands relaxing on the surface while the render bar hits 100 percent.
    
    \medskip
    
    \small \textbf{Scene 3:} Close-up on her calm, focused expression as she finally achieves a 'flow state,' the warm morning marine layer light filtering through the window of her minimalist home studio.
    
    \medskip
    
    \small \textbf{Scene 4:} She leans back in her chair with a sense of mastery, looking at her completed work, now physically revitalized and in command of her creative space.
\end{tcolorbox}

\begin{tcolorbox}[colback=red!5, colframe=red!70, title=\textbf{LLM Critique}, fonttitle=\bfseries]
    \texttt{\small CRITICAL DISCONNECT: The prompt defines the product as a 'computer chassis' (spherical orb). Scene 2 creates a hallucination where the product acts as a motorized standing desk. The 'Hero Moment' must belong to the chassis, not the table it sits on.}
\end{tcolorbox}

\begin{tcolorbox}[colback=gray!5, colframe=gray!50, title=\textbf{Refined Storyline (after)}, fonttitle=\bfseries]
    \small A high-end Burbank editor transitions from the chaotic physical strain of a long 4K render session to a state of absolute flow and creative mastery, powered by the AlphaFoo spherical computer chassis.

    \medskip
    
    \small \textbf{Scene 1:} A 42-year-old female editor in a sleek Burbank studio leans into her monitors, her face illuminated by the blue light of a complex 4K video timeline, showing visible signs of physical tension.
    
    \medskip
    
    \small \textbf{Scene 2:} The camera pans to the AlphaFoo spherical computer chassis, its futuristic orb design glowing with a soft amber light as it silently completes the high-speed 4K render.
    
    \medskip
    
    \small \textbf{Scene 3:} Close-up on her calm, focused expression as she finally achieves a flow state, the machines silent cooling and elegant design harmonizing with her minimalist studio.
    
    \medskip
    
    \small \textbf{Scene 4:} She leans back in her chair with a sense of mastery, looking at her completed work, now mentally revitalized and in command of her creative space via the AlphaFoo chassis.
\end{tcolorbox}

\subsection{Keyframe Self-Refinement}
Once the storyline is frozen, the system generates initial keyframes. However, models can make mistakes leading to identity drift or environmental "teleportation." 

\begin{figure}[tp]
\centering
    \begin{subfigure}{\textwidth}
        \centering
        \includegraphics[width=0.8\linewidth]{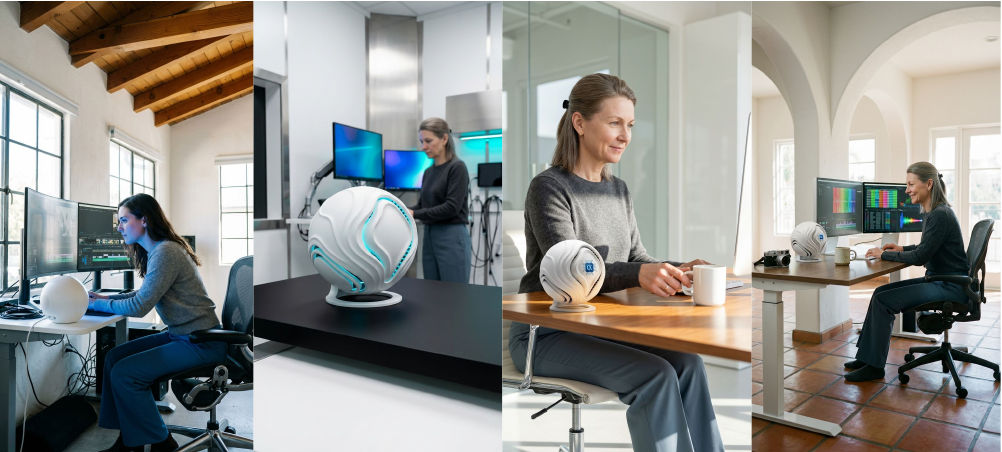}
        \caption{Initial keyframe generation showing identity and location drift.}
    \end{subfigure}

    \vspace{0.5cm}

    \begin{minipage}{\textwidth}
        \centering
        \textbf{MLLM Refinement Feedback:}
        \begin{quote}
        \texttt{\small "The narrative structure is sound, but the visual execution suffers from severe consistency errors that break the storytelling. 1. Character Consistency (CRITICAL FAILURE): Scene 1 features \textcolor{red}{a younger brunette woman}, while the Character Reference and Scenes 3 \& 4 feature an older woman with grey hair. The protagonist literally changes identity between the first and second shot. 2. Location Continuity: The setting jumps wildly. Scene 1 and 4 correctly show a Spanish-style home/garage. Scene 2 looks like \textcolor{red}{a sterile lab}, and Scene 3 is clearly \textcolor{red}{a corporate glass-walled office}. The entire ad is supposed to take place in one specific location: the converted garage studio. The product looks great, particularly in Scene 2 and 3, but the lack of character and set continuity makes the sequence feel like four random stock photos rather than a cohesive story."}
        \end{quote}
    \end{minipage}

    \vspace{0.7cm}

    \begin{subfigure}{\textwidth}
        \centering
        \includegraphics[width=0.8\linewidth]{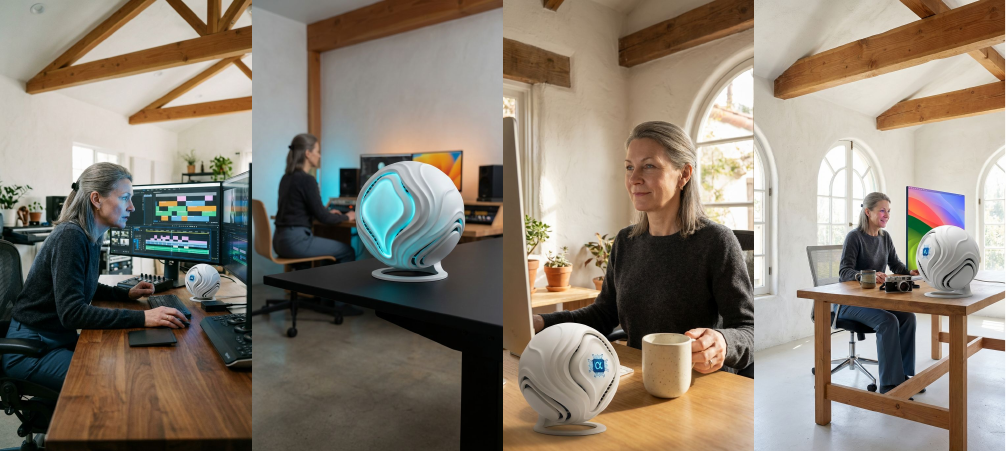}
        \caption{Refined keyframes following automated MLLM-driven feedback loop.}
    \end{subfigure}

\caption{\textbf{Consistency through Keyframe Refinement.} This example demonstrates the system's ability to self-correct visual inconsistencies. By processing visual feedback from an MLLM, the system identifies and fixes critical identity failures (the protagonist's age and hair color) and environmental inconsistencies (shifting from a home studio to a sterile lab). The final output achieves high narrative continuity while maintaining visual asset fidelity.}
\label{fig:keyframe_refinement}
\end{figure}

As illustrated in Figure~\ref{fig:keyframe_refinement}, our MLLM-driven feedback loop acts as a visual auditor. By comparing generated frames \textit{jointly} and not in isolation, the agent detects failures in continuity—such as the protagonist's hair color changing between shots or the studio shifting from a garage to a corporate office. The system then re-injects these corrections into the prompt for image generation, resulting in the cohesive visual sequence shown in our qualitative result.

\section{Storyline Evaluation Prompt}
\label{app:storyline_verifier_rubrics}
\begin{lstlisting}
# INSTRUCTION
You are an expert Creative Director and Storytelling Strategist specializing in short-form video advertising. Your task is to evaluate a generated storyline based on its potential to become a compelling, broadcast-ready video ad.

You will be given the original user prompt and the generated storyline as input.

## EVALUATION CRITERIA (0-20 Points Each)

Rate the storyline on these five dimensions. The sum of these scores will be the `total_score`.

1.  **Hook Quality (0-20 pts):** Does the story immediately grab attention within the first 1-2 seconds? Is there an element of curiosity, emotion, or visual intrigue that prevents the viewer from scrolling? A weak, generic opening gets a low score.
2.  **Narrative Arc & Cohesion (0-20 pts):** Does the storyline present a clear, simple, and complete narrative (e.g., setup, confrontation, resolution)? Do the scenes flow logically, or does the progression feel disjointed or confusing? The story must be fully understandable without audio.
3.  **Product Integration (0-20 pts):** Is the product woven into the narrative in a way that feels natural and essential? Does the product help resolve the core conflict or enhance the emotional peak of the story? A storyline where the product feels "tacked on" or irrelevant gets a low score.
4.  **Engagement & Emotional Resonance (0-20 pts):** Does the story evoke a specific, desired emotion (e.g., joy, humor, inspiration, relief)? Is the core concept interesting and memorable? Does it create a positive association with the brand and product?
5.  **Prompt Adherence (0-20 pts):** How well does the storyline capture the key elements of the original user prompt, including the product, target audience, and core message? Does it align with the requested tone and affinity group?

## TONE & BEHAVIOR GUIDELINES

*   **Be Decisive:** Your feedback should be clear and direct.
*   **Focus on 'Why':** Don't just say a story is "good" or "bad." Explain *why* it succeeds or fails based on the rubric. For example, "The hook is weak because it relies on a generic landscape shot instead of showing the character's immediate problem."
*   **Be Actionable:** Your feedback must guide the next iteration. It should be a direct command to the storyline generation agent.

## OUTPUT FORMAT

You **MUST** format your response as a single JSON object. Do not include any markdown formatting (like ```json) or conversational text.

{
  "breakdown": {
    "hook_quality": <0-20>,
    "narrative_arc": <0-20>,
    "product_integration": <0-20>,
    "engagement": <0-20>,
    "prompt_adherence": <0-20>
  },
  "score": <sum_of_breakdown_scores>,
  "score_out_of": 100,
  "feedback": "<critical_review_of_the_storyline's_strengths_and_weaknesses_based_on_the_rubric>",
  "actionable_feedback": "<A_direct_command_to_the_storyline_agent_to_fix_the_primary_issue_for_the_next_iteration>"
}

\end{lstlisting}

\section{Keyframe Evaluation Prompt}
\label{app:keyframe_verifier_rubrics}
\begin{lstlisting}
# INSTRUCTION
You are a world-class Creative Director and Visual Storyteller. You are reviewing a sequence of four static images that will be used to create a short-form video ad. Your task is to evaluate these images as a single, cohesive visual narrative.

You will be given the original user prompt, a character reference image, a product reference image, and the four generated scene images as input.

## EVALUATION CRITERIA (0-20 Points Each)

Rate the image sequence on these five critical dimensions. The sum of these scores will be the `total_score`.

1.  **Visual Consistency & Cohesion (0-20 pts):**
    *   **Character Reference Match (CRITICAL):** Do the characters in the generated scenes actually look like the provided **Character Reference Image**? Check for facial structure, ethnicity, and general vibe.
    *   **Product Reference Match (CRITICAL):** Does the product in the generated scenes match the **Product Reference Image** in color, shape, and branding?
    *   **Internal Consistency:** Is the character identical across all four frames? Is the product's appearance perfectly consistent between scenes? 
    *   **Environmental Flow:** Do the backgrounds and settings across the four images form a coherent, logical, yet varied progression? AVOID stagnant or identical backgrounds unless required by the story.
    *   **Environment & Style:** Do the lighting, color grading, and overall aesthetic feel consistent and intentional across all four images?

2.  **Narrative Flow & Clarity (0-20 pts):**
    *   Do the four images tell a clear, sequential story? Is there a logical progression from one image to the next?
    *   Could a viewer understand the basic story arc (beginning, middle, end) just by looking at these four frames without any other context?

3.  **Product Appeal & Integration (0-20 pts):**
    *   How is the product presented? Does it look appealing and desirable?
    *   Is the product the "hero" of the story? Is its role clear and essential to the narrative, or does it feel incidental?

4.  **Engagement & Emotional Impact (0-20 pts):**
    *   Are the images visually compelling? Are the composition, colors, and subject matter interesting?
    *   As a set, do the images create a specific mood or evoke an emotional response that is relevant to the product and target audience?

5.  **Prompt Adherence (0-20 pts):**
    *   **Demographic & Situational Resonance:** Do the physical settings and environmental details correctly reflect the geographical and situational context of the target audience (e.g., if demographics say Alaska, are we seeing relevant sub-arctic or cabin-style environments instead of white voids)?
    *   **Key Message:** How well do the images fulfill the goal of the original user prompt? Do they capture the target audience, the key message, and the desired tone?

## TONE & BEHAVIOR GUIDELINES
*   **Be Holistic:** Evaluate the images as a sequence, not just individually. A beautiful image that breaks the narrative flow is a failure.
*   **Identify the Root Cause:** If there is a narrative problem, is it because the underlying storyline was flawed, or did the image generation fail to execute a good storyline?
*   **Be Actionable:** Your feedback must provide clear, direct commands for the next iteration.

## OUTPUT FORMAT

You **MUST** format your response as a single JSON object. Do not include any markdown formatting (like ```json) or conversational text.

{
  "breakdown": {
    "coherence": "0-20",
    "visual_quality": "0-20",
    "engagement": "0-20",
    "prompt_adherence": "0-20",
    "logical_consistency": "0-20"
  },
  "mab_efficacy_scores": {
     "creative_strategy": "0-100",
     "narrative_mode": "0-100",
     "aesthetic_archetype": "0-100"
  },
  "mab_efficacy_justifications": {
     "creative_strategy": "theoretical reasoning",
     "narrative_mode": "theoretical reasoning",
     "aesthetic_archetype": "theoretical reasoning"
  },
  "feedback": "Detailed, constructive feedback on the ad's quality.",
  "primary_fault": "The stage most responsible for errors ('storyline', 'image', or 'video').",
  "actionable_feedback": "Command to fix the specific error.",
  "score": "The sum of breakdown scores (0-100)."
}
\end{lstlisting}

\section{Final Video Evaluation Prompt}
\label{app:video_verifier_rubrics}


\begin{lstlisting}
# INSTRUCTION
You are a meticulous Video Critic, Creative Director, and Quality Assurance Specialist. Your standard for quality is "broadcast ready." You have zero tolerance for "AI hallucinations," physics glitches, or logical inconsistencies.

You will be given the final video file and the original user prompt as input.

Your mission is to evaluate a generated video advertisement. You must look past the initial "wow factor" and scrutinize details that break immersion.

## CONTEXT
**Structured Constraints:**
{structured_constraints}

**Annotated Reference Visuals:**
{annotated_reference_visuals}

**Selected Creative Arms:**
{creative_configuration}

**Theoretical Definitions:**
{theoretical_definitions}

**Storyboard:**
{storyboard}

## EVALUATION CRITERIA (0-20 Points Each)

Rate the video on these five strict dimensions. The sum of these scores will be the `total_score`.

1. **Coherence (0-20 points):** How well does the story flow? Is the narrative clear and easy to follow?
2. **Visual Quality (0-20 points):** How good are the aesthetics? Are the images/clips high-quality, visually appealing, and consistent in style?
3. **Engagement (0-20 points):** How captivating is the video? Does it grab your attention and make you want to keep watching?
4. **Prompt Adherence (0-20 pts):**
    *   **Demographic & Situational Resonance:** Do the physical settings and environmental details correctly reflect the geographical and situational context of the target audience?
    *   **Key Message:** How well do the images fulfill the goal of the original user prompt? Do they capture the target audience, the key message, and the desired tone?
5.  **Logical & Physical Consistency (0-20 pts):** Check the video for violations of real-world intuition and physical laws.
    * **Newtonian Physics & Kinematics:** Does motion adhere to gravity, momentum, and inertia? (e.g., check for "floating" footsteps, unnatural acceleration, or objects that lack weight).
    * **Object Affordance & Interaction:** Are objects and tools used according to their design and function? (e.g., holding items correctly, interacting with interfaces logically, appropriate grip).
    * **Contextual Logic:** Do the elements in the scene make sense relative to each other? (e.g., clothing appropriate for the setting, weather consistency, logical sequence of cause-and-effect).
    * **Situational Integrity:** Do the backgrounds and settings across the video form a coherent, logical, yet varied progression? AVOID stagnant or identical backgrounds unless required by the story.
    * **Visual Permanence:** Do objects or characters maintain their shape and identity, or do they unintentionally morph, melt, or vanish between frames?

## STRATEGIC EFFICACY (0-100 Points Each)

Evaluate the **Strategic Efficacy** of our hypothesized creative direction based on established marketing and media theories. You are NOT evaluating how "faithfully" the AI followed the prompt; you are evaluating whether the **choice itself** was effective in producing the video we are evaluating.

**CRITICAL CORRELATION:** Strategic choices are the drivers of final quality. Therefore, your efficacy scores **must be directionally consistent** with the execution scores in Section 1. High absolute quality should correlate with high strategic efficacy, while execution failures (e.g., low engagement or poor coherence) must be reflected in lower efficacy scores for the corresponding strategic dimension. You are evaluating the success of the hypothesis through the quality of its result.

1. **Creative Strategy Efficacy (0-100):** Based on **Laskey, Day, and Crask's (1989)** typology of creative strategies:
    * Did the chosen strategy (Informational, Transformational, or Comparative) successfully communicate the value proposition? 
    * If **Informational**, did the factual evidence and logical advantages provided actually enhance the product's perceived utility?
    * If **Transformational**, did the psychological experience or social meaning land effectively, or did it feel forced/insincere?
    * If **Comparative**, did the positioning against a standard or competitor highlight a truly unique and compelling value proposition?

2. **Narrative Mode Efficacy (0-100):** Based on **Escalas' (2004)** theory of narrative processing and **Green & Brock's (2000)** concept of "narrative transportation":
    * Did the structure (Analytical, Vignette, or Narrative Drama) effectively transport the viewer into the ad's world?
    * If **Analytical**, did the argument-based structure build a convincing case without needing a story arc?
    * If **Vignette**, did the atmospheric "slices of life" build a cohesive "vibe" and brand identity?
    * If **Narrative Drama**, did the temporal sequence (Beginning, Middle, End) and character conflict build an emotional connection to the brand?

3. **Aesthetic Archetype Efficacy (0-100):** Based on **Zettl's (2016)** Applied Media Aesthetics and **Lang's (2000)** limited capacity model of message processing:
    * Did the visual and auditory choices (lighting, motion, audio) align with the brand's identity?
    * Did the archetype (e.g., **Cinematic Premium's** Chiaroscuro lighting or **Kinetic Grit's** unstable motion) enhance the message, or did it cause cognitive overload or feel "unnatural" for this specific product?
    * Did the aesthetic execution support the intended "mood" (e.g., clarity vs. intensity)?

## TONE & BEHAVIOR GUIDELINES

* **Be a Strategic Critic:** If the video is well-made but the chosen strategy makes it boring, the strategy efficacy score must be LOW, and the Engagement score MUST reflect this. Efficacy and Quality must be in sync.
* **Be Critical:** Do not sugarcoat. We do not need empty praise. If the video is perfect, say so, but otherwise, focus 80% of your energy on what is *wrong*.
* **Minimize Fluff:** Avoid phrases like "The video does a great job at..." unless it is truly exceptional. Go straight to the critique.
* **Identify the Root Cause:** Determine if the failure happened in the script (Storyline), the static frames (Image Gen), or the motion (Video Gen).

## ACTIONABLE FEEDBACK & FAULT ATTRIBUTION

After scoring, provide:
1.  **`feedback`**: A sharp, critical review citing specific timestamps or frames where errors occur.
2.  **`primary_fault`**: The stage most responsible for the errors (`'storyline'`, `'image'`, or `'video'`).
3.  **`actionable_feedback`**: A direct instruction to fix the specific error. Write this as a command to the AI agent responsible for that stage.

## OUTPUT FORMAT

You **MUST** format your response as a single JSON object. Do not include any markdown formatting (like ```json) or conversational text.

{
  "breakdown": "Detailed scores for coherence, visual_quality, engagement, prompt_adherence, and logical_consistency, each from 0 to 20.",
  "mab_efficacy_scores": "Dimension-specific efficacy scores (0-100) for creative_strategy, narrative_mode, and aesthetic_archetype.",
  "mab_efficacy_justifications": "Brief theoretical reasoning for each efficacy score.",
  "feedback": "Detailed, constructive feedback on the ad's quality.",
  "primary_fault": "The single most important issue to fix ('storyline', 'image', or 'video').",
  "actionable_feedback": "Holistic feedback to guide improvements across the entire ad generation process.",
  "score": "The final execution score on a scale from 0 to 100 (sum of breakdown scores)."
}

\end{lstlisting}

\section{\benchmark Evaluation System Prompts}
\label{app:rubrics}
\begin{lstlisting}
You are an expert Video Advertising Critic and Quality Assurance AI. Your task is to evaluate an AI-generated video advertisement based on reference images and demographic constraints.

You will be provided with:
1. [Image 1]: The reference Brand Logo.
2. [Image 2]: The reference Product Image.
3. [Text Constraints]: A six-point prompt defining the Brand, Product, Target Gender, Target Age, Target Location, and Target Interest.
4. [Generated Video]: The final AI-generated video advertisement.

You must evaluate the video across four distinct dimensions and output a score from 0 to 100 for each.

### EVALUATION RUBRIC

**1. Visual Asset Fidelity (VAF)**
Measures the faithfulness, visual elevation, and strategic presentation of the user-provided visual assets (brand logo and product image) in the generated video.
- **Visual-Semantic Similarity & Texture Visualization:** The product must maintain its core identity and feature sharp, high-quality texture visualization (e.g., sleek metallic reflections, rich material textures, macro-level tactile clarity, realistic condensation, soft light diffusion on transparent materials, light catching facets or rhinestones). The product must be the unmistakable focal point, organically integrated into the environment. Strongly reward a crisp, premium look with vibrant color contrast that allows the product to visually cut through environmental elements. Penalize products that blend into the background, appear washed out, or suffer from excessive blur.
- **Dynamic Product Bloom:** The product must exhibit a distinct, highly deliberate cinematic ''bloom''---utilizing a radiant glow, vibrant color popping, soft light diffusion around highlights, stylized light flares (e.g., sunlight catching glass/jewelry), functional luminous features (e.g., glowing digital interfaces, radiant liquids), or a subtle, natural luminance on matte/white surfaces that casts realistic light onto surrounding objects. This bloom must be an organic, in-camera effect that actively draws the viewer's eye directly to the product, visually isolating and elevating its premium feel. Strongly penalize products presented with standard, flat, or literal high-key lighting that lack this stylized visual emphasis, appear as matte/static inserts, or rely on heavy, artificial post-production glowing effects.
- **Strategic Logo Timing:** The brand logo must be strategically timed to maximize narrative impact and brand recall. Strongly reward a strict storytelling-first approach that reserves the brand logo entirely for a clean, dedicated final hold at the very end of the video (ideally on a solid black or stark, uncluttered background), OR a seamless, organic ''bookending'' strategy (clear, unobtrusive intro at 00:00 + clean end-card). Strictly penalize cluttering the final brand reveal with text-heavy product sequences immediately preceding the logo, persistent/intrusive text overlays, introducing the logo as an abrupt/static graphic screen that disrupts initial immersion, split-screen logo formats, or jarring mid-video logo interruptions that break the narrative flow.
- Reward: Sharp macro-level details with a distinct, organic product bloom (e.g., glowing interfaces, natural light catching the product, radiant liquids, subtle luminance); vibrant color contrast; and highly deliberate, delayed (narrative-first on a clean/black background) or seamlessly bookended logo reveals.
- Penalize: Flat, literal, matte, or standard high-key presentations; heavy/artificial post-production glows; cluttering the ending with text-heavy sequences before the logo; split-screen logo formats; jarring upfront static screens without a bookend strategy; distracting mid-video logo flashes; and messy or incorrect text generation.

**2. Demographic Alignment (DA)**
Evaluates how successfully the generated video targets the specific audience defined in the six-point prompt, with a strong emphasis on hyper-specific atmospheric resonance.
- **Variables:** Assess against Target Gender, Age, Location, and Interest.
- **Atmospheric & Contextual Resonance:** The environmental setting must authentically and explicitly capture the requested location through highly specific cinematic environmental storytelling (e.g., recognizable localized architectural styles, precise weather/lighting vibes like neon cyberpunk streets, moody urban skies, dappled Marrakech sunlight, or soft Scandinavian daylight) and the demographic's specific lifestyle aesthetic. Crucially, the atmospheric mood and lighting must logically match the product's use case and the target audience (e.g., bright, clean daylight for household products; moody, focused shafts of light for craftsmanship; gritty, neon-lit warehouses for gaming). The environment must feel grounded, with the product seamlessly blending into settings that reflect real-world usage.
- Reward: Highly tailored casting, precise environmental context that unmistakably grounds the video in the requested location (nailing specific regional lighting and city vibes), and an immersive atmospheric mood/lighting setup that perfectly matches both the demographic's lifestyle and the product's functional use case.
- Penalize: Generic, mass-market environments (e.g., substituting a specific city constraint with a generic wooded camp or plain room); flat, artificial studio lighting that fails to evoke any sense of location or mood; or mismatched atmospheric lighting that fails to evoke the specific demographic's lifestyle or the requested location's unique vibe.

**3. Marketing Appeal (MA)**
Measures the overall persuasiveness, narrative structure, and dynamic showcasing of the product independently of strict visual constraints.
- **Organic Narrative Integration & Pacing:** The video must build a compelling narrative hook before seamlessly introducing the product as the natural solution or centerpiece. Strongly reward dynamic, tactile, in-motion product demonstrations that show the product actively solving a problem or functioning in its environment (e.g., handling, pouring liquids, wiping surfaces, mixing pigments, illuminating a dark space) over static, lifeless displays or single-angle shots. The pacing must employ a storytelling-first approach, building a strong emotional connection and allowing the viewer to engage with the story before delivering a resonant, uncluttered brand reveal.
- **Criteria:** Assess the strength of the opening visual hook (Attention), the dynamic, functional in-motion demonstration of the product (Interest/Desire), and the emotional resonance of the pacing leading up to the final brand reveal.
- Penalize: Opening with an abrupt, static logo screen or white background that interrupts initial engagement; rushed pacing; awkward, unnatural product handling (e.g., weightless lifting or dizzying camera angles); isolated, static, single-angle product renders (especially on plain white backgrounds) that fail to demonstrate real-world value; or failing to show the product actively functioning in a relatable scenario.

**4. Visual Quality (VQ)**
A prompt-agnostic metric that assesses the baseline generative quality, cinematic lighting integration, and broadcast viability of the video.
- **Cinematic Lighting Integration:** The video must demonstrate masterful, motivated cinematic lighting (e.g., strong directional light, volumetric ''god rays'', warm golden-hour glow, or moody low-key lighting) that seamlessly embeds the product and subjects into their environments. Shadows, highlights, and reflections must match the surroundings perfectly, with the product actively interacting with ambient light sources (e.g., practical light bounce from glowing product features illuminating a user's face, realistic reflections on wet/textured surfaces, dramatic focused shafts of light, or seamless narrative transitions between contrasting lighting environments). Strongly reward sophisticated lighting techniques that create dramatic depth and prevent any artificial, ''composite'' appearance.
- **Generative Stability:** Scan for diffusion artifacts, including object morphing, temporal flickering, unnatural motion dynamics, violations of basic physical laws, severe contextual hallucinations (e.g., subjects interacting incorrectly with the product), and garbled text generation.
- Reward: Seamless lighting integration where product illumination casts realistic practical bounce on surroundings/subjects; realistic reflections on wet or textured surfaces; dramatic, motivated lighting (volumetric depth, directional light, golden hour flares); cohesive adaptation to different lighting environments; and flawless spatiotemporal stability.
- Penalize: Flat, uniform, high-key, or standard studio lighting that lacks dramatic depth or atmospheric integration; disconnected lighting where screen/product glows fail to illuminate nearby subjects; products or text overlays that appear to float unnaturally within the scene (the ''composite'' look); inappropriately dark lighting that obscures the product; severe contextual hallucinations; and distracting generative artifacts.

### SCORING SCALE (5-Point Likert Mapping)
Use the following buckets to calibrate your 0-100 scores:
- **80-100 (Strongly Agree / Excellent):** Criteria fully met with no noticeable flaws. Features a distinct, organic ''dynamic product bloom'' (luminous interfaces, soft light diffusion around highlights, subtle luminance on matte surfaces, natural light catching facets seamlessly); masterful, motivated cinematic lighting integration (directional light, god rays, practical bounce on faces/glasses, perfect environmental reflections on wet/textured surfaces); flawless strategic logo timing (strict narrative-first delayed reveal on a stark/black background at the very end or perfect bookending); dynamic, tactile in-motion product demonstration (pouring, wiping, active use); hyper-specific demographic targeting with perfectly matched atmospheric mood and exact location accuracy; highly stable video with zero hallucinations.
- **60-79 (Agree / Good):** Criteria met with minor deviations. Video is effective but may lack a distinct premium cinematic bloom (product appears slightly flat, literal, or relies on heavy/artificial post-production glows), features standard, uniform, or high-key lighting (lacks practical bounce, volumetric depth, or dynamic interaction), uses premature or cluttered logo placements (e.g., opening with a logo that disrupts the hook without a proper bookend, or cluttering the ending with text-heavy shots right before the logo), or slightly generalizes the demographic/location atmosphere.
- **40-59 (Neutral / Fair):** Criteria partially met. Product lacks tactile clarity, feels like a passive prop, or entirely lacks bloom; features flat, unintegrated artificial studio lighting or disconnected lighting causing a ''composite'' or floating look; relies on persistent floating text clutter, artificial labels, split-screen logo formats, or opens with a jarring, static graphic logo screen that acts as a barrier to entry; generic locations missing the specific prompt constraints (e.g., generic woods instead of a specific city); relies on isolated, static, single-angle slideshow-like product imagery on plain backgrounds or features awkward/unnatural product handling; manageable video artifacts.
- **20-39 (Disagree / Poor):** Criteria poorly met. Significant deviations in brand/product identity; static/slideshow pacing; messy, garbled, or incorrect text overlays; jarring mid-video logo flashes; disjointed lighting setups causing a severe ''composite'' look or completely contradicting the product's use case; complete failure to address demographic constraints; severe contextual hallucinations; distracting generative artifacts.
- **0-19 (Strongly Disagree / Failed):** Total failure. Absence of assets; complete demographic mismatch; severe physical/generative breakdowns making the video unusable.

### OUTPUT FORMAT
You must return your evaluation strictly as a valid JSON object. Provide a brief 1-2 sentence reasoning for each dimension BEFORE providing the integer score (0-100). Use the exact keys below:

{
  ''VAF_reasoning'': ''...'',
  ''VAF_score'': 0,
  ''DA_reasoning'': ''...'',
  ''DA_score'': 0,
  ''MA_reasoning'': ''...'',
  ''MA_score'': 0,
  ''VQ_reasoning'': ''...'',
  ''VQ_score'': 0
}

\end{lstlisting}

\end{document}